\documentclass{article} 
\PassOptionsToPackage{numbers}{natbib}
\usepackage{arxiv}

\usepackage{amsmath,amsfonts,bm}

\def\eqref#1{equation~\ref{#1}}

\def\1{\bm{1}}

\DeclareMathAlphabet{\mathsfit}{\encodingdefault}{\sfdefault}{m}{sl}
\SetMathAlphabet{\mathsfit}{bold}{\encodingdefault}{\sfdefault}{bx}{n}

\usepackage{hyperref}
\usepackage{natbib}
\usepackage{arydshln}
\usepackage{adjustbox}
\usepackage[capitalize]{cleveref}
\usepackage{url}
\usepackage{tcolorbox}
\usepackage[ruled,linesnumbered]{algorithm2e}
\usepackage{listings}
\usepackage{multirow}
\usepackage[table]{xcolor}
\usepackage{makecell}
\usepackage{subcaption}
\usepackage{booktabs}
\usepackage{caption}
\usepackage{wrapfig}
\usepackage{tabularx} 
\usepackage{newfloat}
\usepackage{float}
\usepackage{placeins} 
\usepackage{etoc}
\lstdefinestyle{promptbox}{
  basicstyle=\ttfamily\small,
  frame=single,
  framesep=8pt,
  rulecolor=\color{black},
  framerule=0.8pt,
  backgroundcolor=\color{white},
  numbers=none,
  showstringspaces=false,
  showspaces=false,
  showtabs=false,
  breaklines=true,
  breakatwhitespace=true,
  tabsize=2,
  aboveskip=1.2em,
  belowskip=1.2em,
  xleftmargin=0pt,
  xrightmargin=0pt,
  keepspaces=true,
  gobble=0
}

\DeclareFloatingEnvironment[
    fileext=lop,
    listname={List of Prompts},
    name=Prompt
]{prompt}

\DeclareFloatingEnvironment[
    fileext=loi,
    listname={List of Insights},
    name=Insight
]{insight}

\DeclareFloatingEnvironment[
    fileext=lor,
    listname={List of Reasoning Processes},
    name=Raw Reasoning Process
]{reasoningprocess}

\DeclareFloatingEnvironment[
    fileext=lor,
    listname={List of Reasoning Traces},
    name=Reasoning Trace
]{reasoningtrace}

\crefname{prompt}{Prompt}{Prompts}
\Crefname{prompt}{Prompt}{Prompts}

\crefname{insight}{Insight}{Insights}
\Crefname{insight}{Insight}{Insights}

\crefname{reasoningprocess}{Raw Reasoning Process}{Raw Reasoning Processes}
\Crefname{reasoningprocess}{Raw Reasoning Process}{Raw Reasoning Processes}

\crefname{reasoningtrace}{Reasoning Trace}{Reasoning Traces}
\Crefname{reasoningtrace}{Reasoning Trace}{Reasoning Traces}

\tcbset{
  colback=white,
  colframe=cyan!90!blue,
  arc=3mm,
  boxrule=0.8pt,
  left=6pt,
  right=6pt,
  top=6pt,
  bottom=6pt,
  toptitle=1pt,
  bottomtitle=1pt,
  lefttitle=6pt,
  fontupper=\fontsize{6.5}{8.5}\selectfont\ttfamily\raggedright
}

\title{Federation over Text: Insight Sharing for Multi-Agent Reasoning}

\author{Dixi Yao\footnotemark[0] \footnotemark[2] \thanks{Correspondence
to: Dixi Yao $<$dixi@uchicago.edu$>$} \quad
Tahseen Rabbani\footnotemark[2] \quad
Manzil Zaheer\footnotemark[3] \quad
Tian Li  \footnotemark[2] \\
 \footnotemark[2] University of Chicago 
 \quad
\footnotemark[3]~ Google DeepMind
}
\date{}

\begin{document}

\maketitle

\etocdepthtag.toc{mtchapter}

\begin{abstract}
Modern agents specialize in varying domains while there is no clear approach combining different domain skills. We propose a federated learning-like framework, \emph{Federation over Text} (FoT), that enables multiple clients solving different tasks to collectively generate a shared library of metacognitive insights by iteratively federating their local reasoning processes without sharing actual problem instances. Instead of federation over gradients (e.g., as in distributed training), FoT operates at the \textbf{semantic level} without any gradient optimization or supervision signal. 
At each round, client LLM agents independently apply arbitrary local reasoning and self-improvement procedures to their own tasks and share the resulting reasoning traces with a central server. The server then aggregates, distills, and consolidates knowledge across tasks and domains into a shared insight library, which can be reused by current and future agents to improve their reasoning.
Experiments show that FoT improves reasoning effectiveness and efficiency across real-world daily tasks, cross-domain collaboration, and research insight discovery, achieving an average performance gain of 11.9 absolute percentage points while reducing client-generated task-completion tokens by 5.5\%.
\end{abstract}

\section{Introduction}
LLM-powered multi-agents are increasingly deployed for complex, multi-step tasks~\cite[e.g.,][]{yoran2023answering,liang2024encouraging,lu2024ai,schmidgall2025agent,yu2025researchtown,zhang2026vibe}, including recently solving the Navier--Stokes problem~\citep{openai2026navier}.  However, they can suffer from two major limitations. First, reasoning can be inefficient, as agents often reason from scratch given an input instance, repeatedly solving similar problems, which can be computationally expensive and slow. Second, reasoning in multi-agent systems typically occurs in an isolated manner, where agents are chained by passing prompts or intermediate outputs~\cite[e.g.,][]{ke2025survey}. While agents can exchange task solutions within a session, their internal thinking processes and full reasoning histories are not globally stored or communicated between one another. Hence, hard-won thinking processes critical to a solution cannot be reused or shared across agents.


It remains underexplored how to automatically transfer, reuse, or distill independent reasoning traces to help guide and improve future reasoning processes for similar agents or across domains.
At the same time, classical federated paradigms for knowledge sharing~\citep{mcmahan2017communication} may not be suitable for learning generalizable insights from agents' reasoning traces. Averaging the gradients or weights of entire (or parts of) large generative models can be computationally prohibitive, impossible for closed-source LLMs, and, more importantly, not interpretable, as parameter aggregation is not explicitly equivalent to sharing \textit{abstract, high-level reasoning summaries}~\citep{cao2026empirical}. 

To this end, we propose \emph{federation over text} (FoT), a federated-like framework that enables a set of agents to collaboratively and iteratively build a shared, evolving library of reusable \textit{insights} from independent local reasoning traces (plaintext) without communicating actual problem instances.
Unlike multi-agent systems that decompose a single complex task into subtasks, FoT operates across multiple independent agents on diverse but related tasks.

Unlike traditional distributed learning, which averages gradients or model updates in a high-dimensional, uninterpretable space, FoT operates at the \textbf{semantic}, text level without any parameter optimization, fine-tuning, or reinforcement learning. 
Operating in the semantic text space also provides more interpretable control over what information is transmitted.
In particular, at each round, 
each agent can run an arbitrary local reasoning and self-improvement procedure without requiring external labels or verifiers, and then transmit concise metacognitive summaries of its problem-solving experience to a server.
Instead of simply pooling agents' solutions or traces, the server identifies and distills recurring reasoning fragments from these summaries into new, explicit insights and sends the insight library back to the agents for the next round.  We note that FoT is a \textit{general} framework that allows each agent to use any local model and self-improvement workflow for its individual tasks, as long as summarized reasoning traces are shared. A detailed comparison between FoT and other multi-agent knowledge sharing baselines is in Table~\ref{tab:evolution-comparison} and Section~\ref{sec:applications}.

We apply FoT to various applications including real-world daily tasks, multi-domain collaboration, and research insight discovery (Section~\ref{sec:applications}), and compare FoT with isolated reasoning and other cross-agent knowledge-sharing baselines. 
In real-world daily tasks and multi-domain collaboration, FoT improves performance by 11.9 points on average while reducing client-generated task-completion tokens by 5.5\% relative to isolated reasoning. 
For research insight discovery, insights aggregated from ICLR 2025 papers align with the core technical contributions of 26.93\% more ICLR 2026 papers than insights generated directly from the LLM. FoT enables LLM-based systems to function not only as performance-driven solvers, but also as tools for decision support and insight discovery.

\section{Related Work}

\textbf{Collaborative and Federated Learning.}
Collaborative or federated learning (FL) \citep{mcmahan2017communication} aggregates model parameters from multiple clients while keeping data local. Existing literature on federated agents primarily focuses on training language models in the classical parameter or gradient-level aggregation framework \citep{li2025fedcot}, which can be expensive and insufficient for capturing high-level abstract reasoning strategies. Social learning \citep{mohtashami2023social} allows teacher agents to generate synthetic examples or abstract prompts that are aggregated to train a student model in a single-task setting.
FL and FoT share interesting connections; FoT operates in the semantic space while being compatible with heterogeneous agent models and task domains.

\textbf{LLM Skills and Recursive Self-Improvement.}
To incorporate past experience into LLM agents, existing approaches such as Anthropic Skills~\citep{anthropic_skills}, 
Vibe Research~\citep{zhang2026vibe}, and Scientific Skills~\citep{claudescientificskills2026} primarily provide concrete procedural knowledge for specific tasks, rather than generalizable and abstract insights. Many works rely on human-provided supervision or external evaluation (e.g., labels, validation sets, or execution feedback) to identify successful behaviors and refine memories or skills~\citep{zhao2024expel,zhang2025agentic,qu2026coral,yang2026federatedskill,zhang2026hyperagents,yin2025godel,zhuang2026test,wang2025evolving,zweigerself,yoran2023answering,wang2024agent,rao2026skillopt}. FoT allows the use of any local self-evolution method, without necessarily requiring any external validation or labels. 
Other methods learn from self-generated trajectories, but often remain coupled to particular reasoning procedures~\citep{didolkar2025metacognitive,ye2026meta,yang2024buffer,wang2024agent}. FoT is designed to be agnostic to specific local algorithms, where the reasoning traces are aggregated into generalizable insights. This differs from methods that prescribe specific local pipelines for extracting and updating skills~\citep{yang2024buffer, zhao2024expel} (\Cref{sec:properties}).

\textbf{Multi-Agent Collaboration.}
Extensive prior work on multi-agent systems focuses on enabling agents to \textit{solve a given complicated task} collaboratively. A common paradigm involves decomposing a task into smaller subtasks, assigning each to an individual agent, and aggregating the partial results \citep[e.g.,][]{zhang2024chain,tran2025multi,dangmulti,chung2025station,lyu2026evoscientist}. Other existing approaches include multi-agent debate \citep{liang2024encouraging} or allowing multiple agents to repeatedly perform the same task and vote on a solution \citep{wu2024shall,zhang2024more, zhengthought}. 
Previous works also study large-scale agentic knowledge exchange by building concrete workflows for agent laboratories \citep{schmidgall2025agent}, or simulating research communities~\citep{yu2025researchtown,zhang2025aixiv}. 
There are recent works~\citep{qu2026coral,yang2026federatedskill} that propose to pool together reasoning processes from multiple agents. However, they require all agents to solve the same types of tasks, need extra validation signals, and do not directly support flexible and arbitrary local
reasoning methods; see empirical comparisons in Section~\ref{sec:applications}. 

\vspace{-.5em}
\section{Federation over Text}
\vspace{-.5em}

In this section, we propose our federation over text framework, motivated by its connections with federated learning (i.e., gradient aggregation)~\citep{mcmahan2017communication}.
Federated learning (FL) focuses on training a shared global model to serve existing and future clients, whereas in text space we aim to build an {insight library} that helps both existing and new agents better solve a variety of tasks. In the most common FL workflow, the server broadcasts the current global model to each client, which then performs local training on its local data and transmits the local model updates back to the server. The server aggregates the model updates to form a new global model. This iterative process repeats until convergence.  

\begin{figure}[!htbp]
	\centering
	\includegraphics[width=\linewidth]{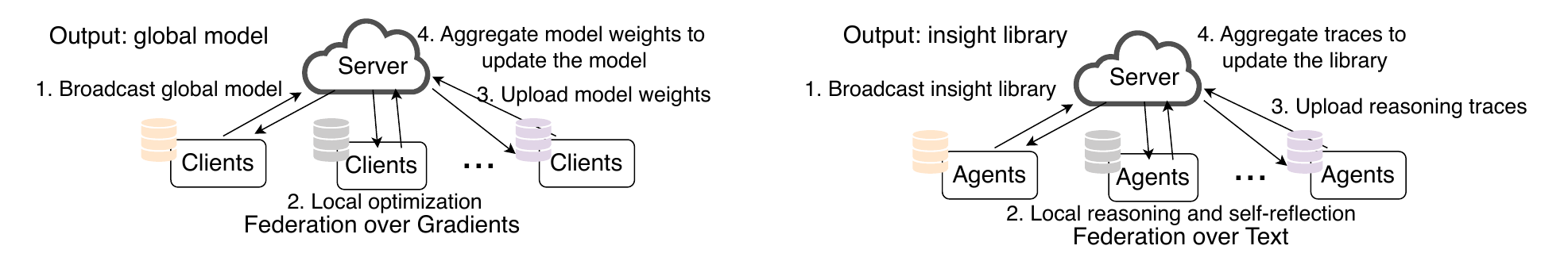}
    \captionsetup{skip=2pt}
	\caption{Workflow of federation over (accumulated) gradients and federation over text. See pseudocode in Algorithms~\ref{alg:fedavg} and \ref{alg:fot} in \Cref{{sec:alg}}.
		}
	\label{fig:workflow}
    \vspace{-0.5em}
\end{figure}

At a high level, in FoT, at each round, each client runs a local agent (any base LLM(s)) that solves its local task with the help of the current insight library, and sends the reasoning traces to the server. The server merges them to update the library and distributes it to all agents for the next round. We present the workflow of FoT in \Cref{fig:workflow} and the pseudocode of both algorithms in Appendix~\ref{sec:alg}.

\textbf{Local Reasoning.} In the workflow of FoT (\Cref{fig:workflow}, right), each of the  decentralized agents solves a specific {problem} or task using a selected base LLM. These tasks can span domains such as mathematical reasoning, coding, and scientific question-answering. Given a task (e.g., a collection of math questions), an agent first solves the task, then reflects on whether the reasoning process contains reusable techniques and what key skills can be extracted. 
Instead of uploading complete, raw reasoning histories, each agent generates a logic-level \textit{reasoning trace} for each problem based on its solution and reflections, specifying when and how to apply the skills via step-by-step instructions. Agents then share these shorter, distilled messages with the server, while keeping problem instances and raw trajectories local. {We use \Cref{prompt:reflection} and \Cref{prompt:insight} (in Appendix~\ref{prompt:solution}) as general guidelines for self-evolution and reasoning-trace construction. In our experiments (\Cref{sec:applications}), we instantiate FoT with several local reasoning methods.}



\textbf{Global Aggregation.} In each round, after collecting traces, the server aggregates them to update the insight library. Unlike prior approaches that match each new trace to a single existing skill, our aggregation allows each trace or insight to contribute to multiple candidate insights. This many-to-many structure is important because a single trace may contain knowledge relevant to multiple domains, while traces from different domains may share an underlying principle despite low surface-level similarity. In particular, the server {reasons at the textual level and first clusters} reasoning traces and past insights that share similar skills, then identifies connections among traces that solve the same problem using different skills and can potentially be merged. {A key property enabling cross-domain merging is that each trace or insight can contribute to multiple potential new insights.}
Finally, the server broadcasts the updated insight library to all agents for the next iteration. The aggregated library can be reused as an external reference to improve the reasoning of existing or future agents outside the current network. Unlike parameter updates, the resulting knowledge is human-readable and interpretable. This improves transparency and enables cross-domain transfer of reasoning strategies. We provide all designed prompts in Appendix~\ref{sec:prompt}. FoT is modular in that local or global procedures can adopt various algorithms/prompts (Section~\ref{sec:ablation}). 

\begin{figure}[b]
\centering
\vspace{-1em}
\includegraphics[width=\linewidth]{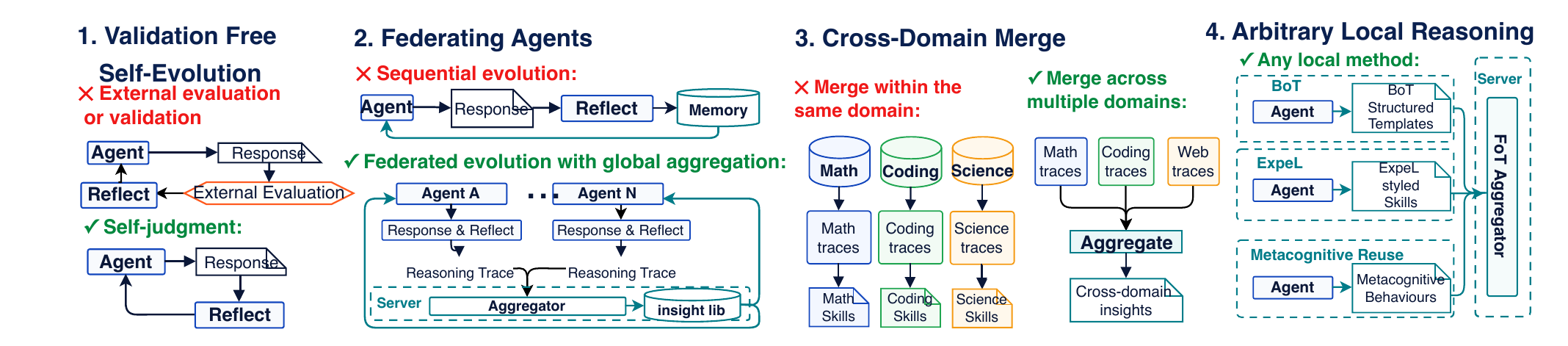}
\captionsetup{skip=2pt}
\caption{Illustration of FoT features; comparison with prior work is in Table~\ref{tab:evolution-comparison}.}
\label{fig:diff}
\vspace{-1em}
\end{figure}

\begin{wraptable}{r}{0.49\linewidth}
    \centering
    \vspace{-1.2em}
    \caption{
        Comparison of experience-utilization methods.
        \Cref{fig:diff} illustrates the distinction between the presence and absence of each feature.
    }
    \label{tab:evolution-comparison}
    \scriptsize
    \setlength{\tabcolsep}{1pt}
    \renewcommand{\arraystretch}{1.12}
    \resizebox{\linewidth}{!}{%
    \begin{tabular}{@{}lcccc@{}}
        \toprule
        \textbf{Method}
        & \makecell{\textbf{Validation}\\\textbf{Free}}
        & \makecell{\textbf{Federating}\\\textbf{Agents}}
        & \makecell{\textbf{Cross-Domain}\\\textbf{Merge}}
        & \makecell{\textbf{Arbitrary Local}\\\textbf{Reasoning}} \\
        \midrule

        ExpeL ~\citep{zhao2024expel}
        & $\times$ & $\times$ & $\checkmark$ & -- \\

        BoT ~\citep{yang2024buffer}
        & $\checkmark$ & $\times$ & $\times$ & -- \\

        AWM ~\citep{wang2024agent}
        & $\times$ & $\times$ & $\times$ & -- \\

        \makecell[l]{Metacognitive Reuse\\~\citep{didolkar2025metacognitive}}
        & $\checkmark$ & $\times$ & $\times$ & -- \\

        HyperAgents ~\citep{zhang2026hyperagents}
        & $\times$ & $\times$ & $\checkmark$ & -- \\

        CORAL ~\citep{qu2026coral}
        & $\times$ & $\checkmark$ & $\times$ & $\times$ \\

        FederatedSkills ~\citep{yang2026federatedskill}
        & $\times$ & $\checkmark$ & $\times$ & $\times$ \\
        \hline
        FoT (ours)
        & $\checkmark$ & $\checkmark$ & $\checkmark$ & $\checkmark$ \\

        \bottomrule
    \end{tabular}%
    }
        \vspace{-2em}
\end{wraptable}

\textbf{Difference from Existing Experience-Reuse Frameworks.}  The workflow above resembles several existing approaches that use past agent experience to improve future reasoning, but FoT differs in how experience is generated, shared, and aggregated. We summarize these distinctions through four properties in Figure~\ref{fig:diff} and Table~\ref{tab:evolution-comparison}.
\textit{Validation-free self-evolution} extracts reusable knowledge through self-reflection without external labels, rewards, test cases, or execution feedback. \textit{Federated agents} independently solve local tasks and contribute traces to a shared server that actively synthesizes knowledge across agents, rather than sequentially updating or simply pooling memory. \textit{Cross-domain merge} combines traces from potentially different domains into shared higher-level insights, instead of only matching or consolidating similar within-domain experiences. Finally, \textit{compatibility with arbitrary local reasoning} decouples server-side aggregation from the client-side reasoning procedure, allowing FoT to incorporate different models within the same framework.

\section{Applications of FoT} \label{sec:applications}
In \Cref{sec:openclaw}, we study real-world daily tasks spanning web research, communication, and productivity, where agents can benefit from sharing reusable experience across related tasks.
In \Cref{sec:mix3analysis}, we apply FoT to cross-domain collaboration including mathematics, science, and coding, where we evaluate whether insights can transfer to new domains. We additionally demonstrate that FoT can distill shared technical insights from machine learning papers in Section~\ref{sec:insight}.
Our code and experiments are open sourced at \url{https://github.com/dixiyao/FoT}. 

We compare FoT with several forms of experience reuse and agent collaboration. `Isolated agents' solve each task independently without shared experience, while `RAG of raw traces' stores previous reasoning traces and retrieves them directly when solving new tasks.
We also evaluate existing sequential self-evolution methods, including Metacognitive Reuse~\citep{didolkar2025metacognitive}, HyperAgents~\citep{zhang2026hyperagents}, ExpeL~\citep{zhao2024expel}, AWM~\citep{wang2024agent}, and BoT~\citep{yang2024buffer}, and federated methods CORAL~\citep{qu2026coral} and FederatedSkills~\citep{yang2026federatedskill}. As FoT is a modular and flexible framework, we also evaluate the performance of plugging local reasoning strategies (Metacognitive Reuse, HyperAgents, ExpeL, and AWM) into FoT. Finally, to test different global aggregation approaches, we compare FoT aggregation with pooling, RAG, Context Compaction~\citep{anthropic2024claudecookbooks}, and Chain-of-Density (CoD)~\citep{adams2023sparse}, as well as variants that remove inter-agent sharing or the trace relationship construction step (Prompt~\ref{prompt:relation}).

\subsection{Real-World Daily Tasks}
\label{sec:openclaw}
We first evaluate FoT on general-purpose agent tasks using OpenClaw~\citep{openclaw2025openclaw} (2026.6.33) as the agent framework. We use PinchBench~\citep{pinchbench} and ClawEval~\citep{ye2026claw}, which cover tasks such as web research, financial analysis, software development, professional communication, content creation, and productivity workflows. Each agent is assigned an individual task and produces a reasoning trace from its execution trajectory.

We first run the methods on the 23 tasks in PinchBench V1\footnote{The initial 23 tasks released by PinchBench.}, from which shared experience or insight libraries are constructed. We then reuse the resulting shared knowledge on held-out PinchBench tasks and on ClawEval, allowing us to distinguish improvement on tasks participating in federation from generalization to unseen tasks. For self-evolution methods (FoT, BoT, and Metacognitive Reuse), no labels or execution results are provided, while the other methods (HyperAgents, ExpeL, AWM, CORAL, and FederatedSkills) receive labels, rewards, or program execution results according to the details of each algorithm. 
For methods using shared memories, we let them use the memories to solve held-out tasks. We have an additional baseline where each agent is equipped
with a set of human-defined skills provided by OpenClaw (``Curated Skills''). 


We report our main results in \Cref{tab:pinchbench_results}. FoT improves reasoning effectiveness over the baselines on both existing and unseen tasks. Semantic aggregation (rather than simply sharing more experience) is crucial for the gains: across multiple local reasoning strategies, replacing pooling with FoT aggregation consistently improves held-out performance. Additionally, aggregating only within each individual agent substantially reduces held-out performance, showing the benefit of sharing experience across agents.
In terms of efficiency, FoT shares distilled reasoning traces and compact insight libraries rather than complete raw LLM outputs. For instance, under four local reasoning strategies, FoT reduces shared context by approximately 23K tokens on average relative to pooling the traces from those reasoning methods. 
\Cref{sec:cost} provides the corresponding monetary-cost analysis.

\newcommand{\graycell}[1]{\cellcolor{gray!8}#1}
\begin{table*}[!t]
\vspace{-0.5em}
    \centering
    \caption{
    Performance and efficiency on real-world daily tasks. We compare different local reasoning strategies and global aggregation approaches, together with non-evolving and federated baselines. `Shared' denotes the number of tokens in the shared knowledge, `Extra Tokens' denotes additional tokens for client-side evolution and server-side aggregation, and \# G.T. denotes client-generated tokens for task completion. Results are the mean ($\pm$ std) over three independent runs. DeepSeek-V4-Flash-0731 is the base model, and GPT-Luna-5.6-Pro is used when LLM-as-a-judge is required. We report results of agents running for 5 iterations.
    }
    \label{tab:pinchbench_results}
    \footnotesize
    \setlength{\tabcolsep}{2.3pt}
    \renewcommand{\arraystretch}{1.12}

    \begin{adjustbox}{max width=\textwidth}
\begin{tabular}{@{}llll*{6}{c}@{}}
    \toprule
     &
     &
    &  &
    \multicolumn{2}{c}{
        \textbf{\makecell{Existing Tasks\\PinchBench V1}}
    } &
    \multicolumn{2}{c}{
        \textbf{\makecell{Held-Out Tasks\\PinchBench V2 $\setminus$ V1}}
    } &
    \multicolumn{2}{c}{
        \textbf{\makecell{New Tasks\\ClawEval}}
    } \\
    \cmidrule(lr){5-6}
    \cmidrule(lr){7-8}
    \cmidrule(lr){9-10}

     \textbf{Method} & \textbf{Variant} & \makecell{\textbf{Shared (K)}} & \makecell{\textbf{Extra }\\\textbf{Tokens (K)}}
     &
    \textbf{Success Rate} &
    \makecell{\textbf{\# G.T.}\\\textbf{(K)}} &
    \textbf{Success Rate} &
    \makecell{\textbf{\# G.T.}\\\textbf{(K)}} &
    \textbf{Score} &
    \makecell{\textbf{\# G.T.}\\\textbf{(M)}} \\
    \midrule

    \rowcolor{gray!36}
    \multicolumn{10}{@{}l}{
        \textbf{Non-Evolving Baselines}
    } \\

    \multirow{3}{*}{
        \makecell{ Isolated Agents\\(ReAct)}
    }
    &  -- 
    &  --
    & --
    & 84.06 (5.42)
    & \textbf{41 (1)}
    & 84.68 (1.32)
    & 818 (17)
    & 56.52 (0.51)
    & 1.52 (1.12) \\

    &  + Curated Skills
    & 6
    & --
    & 85.51 (2.05)
    & 70 (7)
    & 86.29 (1.32) 
    & 838 (11)
    & 61.90 (0.29)
    & 1.08 (0.83) \\

    &  + RAG of raw traces
    & 4
    & --
    & 86.96 (0.00)
    & 54 (3)
    & 62.37 (6.76)
    & 864 (53)
    & 45.76 (0.77)
    & 1.23 (1.54) \\

    \midrule
    \rowcolor{gray!36}
    \multicolumn{10}{@{}l}{
        \textbf{Sequential Evolution Methods}
    } \\

    \multirow{3}{*}{
        \makecell[l]{
            Metacognitive Reuse~\citep{didolkar2025metacognitive}
        }
    }
    &  --
    & --
    & \multirow{3}{*}{111 (4)}
    & 86.96 (0.00)
    & 57 (3)
    & --
    & --
    & --
    & -- \\

    &  + Pooling
    & 39 (7)
    &
    & 89.86 (2.05)
    & 51 (2)
    & 76.08 (1.66)
    & 790 (75)
    & 70.19 (0.51)
    & 1.52 (1.39)\\

    &  + RAG
    & 4
    &
    & 82.61 (3.55)
    & 57 (7)
    & 73.39 (2.28)
    & 837 (97)
    & 60.04 (1.06)
    & 5.67 (2.85)\\

    \multirow{3}{*}{
         HyperAgents~\citep{zhang2026hyperagents}
    }
    &  --
    & --
    & \multirow{3}{*}{58 (3)}
    & 89.86 (2.05)
    & 66 (6)
    & --
    & --
    & --
    & -- \\

    & + Pooling
    & 22 (5)
    &
    & 89.86 (2.05)
    & 62 (2)
    & 82.26 (0.66)
    & 913 (56)
    & 68.32 (0.51)
    & 1.44 (0.89)\\

    &  + RAG
    & 4
    &
    & 89.86 (2.05)
    & 48 (1)
    & 79.03 (1.32)
    & 873 (87)
    & 60.66 (0.77)
    & 1.32 (0.76)\\

    \cdashline{1-10}

    ExpeL~\citep{zhao2024expel}
    & -
    & 28 (3)
    & 52 (2)
    & 88.41 (2.05)
    & 59 (4)
    & 83.60 (2.66)
    & \textbf{732 (64)}
    & 69.15 (0.29)
    & \textbf{1.02 (0.93)} \\

    AWM~\citep{wang2024agent} (online)
    &  -
    & 56 (2)
    & 56 (2)
    & 88.41 (2.05)
    & 58 (5)
    & 71.24 (2.66)
    & 997 (24)
    & 67.29 (0.77)
    & 1.26 (0.83) \\

     BoT~\citep{yang2024buffer}
    &  -
    & 118 (12)
    & 353 (53)
    & 91.30 (0.00)
    & \textbf{41 (4)}
    & 86.56 (1.37)
    & 970 (32)
    & 62.53 (1.06)
    & 1.73 (1.23)\\

    \midrule
    \rowcolor{gray!36}
    \multicolumn{10}{@{}l}{
        \textbf{Federated Evolution Methods}
    } \\

     CORAL~\citep{qu2026coral}
    &  -
    & 56 (3)
    & 43 (2)
    & 88.41 (2.05)
    & 50 (5)
    & 74.46 (1.66)
    & 832 (18)
    & 70.81 (0.51)
    & 2.11 (1.67)
    \\

     FederatedSkills~\citep{yang2026federatedskill}
    &  -
    & 21 (2)
    & \textbf{32 (1)}
    & 88.41 (2.05)
    & 45 (8)
    & 76.08 (1.01)
    & 798 (55)
    & --
    & --
    \\

    \midrule
    \textbf{Local} & \textbf{Server} & &  & & \\
    \midrule
    \rowcolor{gray!36}
    \multicolumn{10}{@{}l}{
        \textbf{Various Local-Reasoning Strategies  Plugged into FoT}
    } \\
    
    Metacognitive Reuse~\citep{didolkar2025metacognitive}
    &  \multirow{4}{*}{ FoT Aggregation} 
    & 3 (1)
    & 127 (5)
    & 92.75 (2.05)
    & 52 (10)
    & 90.86 (2.66)
    & 907 (76)
    & 74.33 (0.29)
    & 4.97 (1.78)\\
    
     HyperAgents~\citep{zhang2026hyperagents}
    &  
    & 4 (1)
    & 73 (2)
    & 92.75 (2.05)
    & 55 (7)
    & \textbf{92.20 (1.01)}
    & 989 (63)
    & \textbf{75.98 (1.17)}
    & 1.54 (0.82)\\

   ExpeL~\citep{zhao2024expel}
    & 
    & 4 (1)
    & 72 (3)
    & 89.86 (4.10)
    & 57 (6)
    & 91.67 (1.01)
    & 930 (70)
    & 73.29 (0.88)
    & 1.70 (1.28)\\

    AWM~\citep{wang2024agent}
    & 
    & 5 (1)
    & 72 (2)
    & 92.75 (2.05)
    & 51 (2)
    & \textbf{92.20 (1.52)}
    & 915 (26)
    & 74.12 (1.28)
    & 1.51 (0.89)\\
    
    \midrule
    \rowcolor{gray!36}
    \multicolumn{10}{@{}l}{
        \textbf{FoT with Various Global Aggregation Strategies}
    } \\

    \multirow{7}{*}{ FoT Local}
    &  FoT Aggregation on each agent
    & 40 (5)
    & 90 (6)
    & 89.86 (2.05)
    & 49 (5)
    & 77.42 (3.95)
    & 863 (88)
    & 73.71 (1.06)
    & 1.97 (0.96) \\

    \cdashline{2-10}
    
    &  Pooling
    & 28 (3)
    & \multirow{4}{*}{54 (2)}
    & 86.96 (0.00)
    & 59 (4)
    & 76.88 (1.52)
    & 793 (86)
    & 69.57 (1.34)
    & 1.98 (1.58)\\

    &  RAG
    & 4
    &
    & 85.51 (2.05)
    & 46 (2)
    & 75.81 (2.28)
    & 810 (56)
    & 62.53 (1.17)
    & 1.45 (0.87) \\
    
    & Compaction~\citep{anthropic2024claudecookbooks}
    & 10 (1)
    &
    & 89.86 (2.05)
    & 43 (5)
    & 75.54 (2.74)
    & 841 (21)
    & 72.05 (0.51)
    & 1.45 (1.07) \\

    &  CoD~\citep{adams2023sparse}
    & \textbf{0.2}
    &
    & 89.86 (2.05)
    & \textbf{41 (3)}
    & 67.74 (1.74)
    & 763 (63)
    & 63.98 (1.01)
    & 1.61 (0.71)\\

    \cdashline{2-10}

    &  FoT Aggregation (w/o \Cref{prompt:relation})
    & 2 (1)
    & 60 (3)
    & 91.30 (3.55)
    & 48 (3)
    & 83.60 (4.28)
    & 838 (29)
    & 68.74 (3.60)
    & 1.52 (1.11)\\

    &  FoT Aggregation
    & 2 (1)
    & 66 (4)
    & \textbf{94.20 (2.05)}
    & 49 (2)
    & \textbf{92.20 (0.76)}
    & 938 (28)
    & 74.74 (0.29)
    & 1.31 (1.11) \\

    \bottomrule
\end{tabular}
    \end{adjustbox}
    \begin{minipage}{0.99\textwidth}
        \footnotesize
    1. RAG has a 4,096-token budget per retrieval because FoT Aggregation generates libraries of 2 to 5K tokens.
        
    2. For pooling, Shared counts all tokens in the collected traces or experiences.

    3. FederatedSkills creates one skill cluster for each type of task. Because PinchBench and ClawEval have different task categories and distributions, FederatedSkills cannot be applied to held-out ClawEval.
    \end{minipage}
    \vspace{-1.5em}
\end{table*}

\subsection{Multi-Domain Collaboration}
\label{sec:mix3analysis}
In this section, we investigate how FoT retrieves cross-domain knowledge, helping agents working on different but related domain tasks collaborate and improve each other's reasoning. We construct the insight library from Humanity's Last Exam (HLE)~\citep{phan2025humanity}, which includes tasks over ten different domains, including science, AI, engineering, etc. We use 2,500 agents to generate reasoning traces across these heterogeneous tasks. 
The resulting library is then evaluated both on HLE and on three held-out benchmarks from different domains: AIME26~\citep{aime26}, GPQA Diamond~\citep{reingpqa}, and LiveCodeBench V6~\citep{jainlivecodebench}.

\Cref{tab:multi-domain} shows that FoT improves performance on both the tasks used to construct the library and the held-out domains, indicating that the distilled insights also transfer to different task distributions and domains. Simple sharing mechanisms (such as trace pooling and RAG) are substantially less effective on both existing and unseen tasks. Replacing our proposed FoT aggregation with other global baselines (pooling, Compaction, or CoD) also reduces performance, showing the superiority of FoT aggregation over simply storing or retrieving the traces independently.


To further explore the benefits of cross-domain aggregation, we let each agent mark which insights contribute most to the reasoning process. Agents frequently select insights synthesized from traces originating in different domains, and some insights are reused across multiple task categories. In \Cref{sec:supplymultidomain}, we present \textbf{two case studies} illustrating the content of the generated insights, which insights are used, and how they are applied during the reasoning process to arrive at correct solutions. These examples show that FoT can generate insights from multiple domains (e.g., math, CS) and use them to correct the reasoning processes of tasks in a different domain (e.g., chemistry). 
We provide two concrete examples comparing agents with and without the insight library in \Cref{apx:toolcallexample}. 

\begin{table*}[!htbp]
    \centering
    \captionsetup{skip=2pt}
    \caption{
    Performance and efficiency on multi-domain collaboration. Gemini 2.5 Flash Lite is the base model. We run for 3 iterations. All other settings are the same as in \Cref{tab:pinchbench_results}.}
    \label{tab:multi-domain}
\captionsetup{skip=2pt}
\footnotesize
\setlength{\tabcolsep}{2.3pt}
\renewcommand{\arraystretch}{1.12}
\begin{adjustbox}{max width=\textwidth}
\begin{tabular}{@{}ll*{8}{c}@{}}
    \toprule
    &
    &
    \multicolumn{2}{c}{
        \textbf{\makecell{Existing Task\\HLE}}
    } &
    \multicolumn{2}{c}{
        \textbf{\makecell{Held-Out Task\\AIME26}}
    } &
    \multicolumn{2}{c}{
        \textbf{\makecell{Held-Out Task\\GPQA Diamond}}
    } &
    \multicolumn{2}{c}{
        \textbf{\makecell{Held-Out Task\\LiveCodeBench V6}}
    } \\
    \cmidrule(lr){3-4}
    \cmidrule(lr){5-6}
    \cmidrule(lr){7-8}
    \cmidrule(lr){9-10}

    \textbf{Method} &
    \textbf{Variant} &
    \textbf{Acc. (\%)} &
    \makecell{\textbf{\# G.T. (M)}} &
    \textbf{Acc. (\%)} &
    \makecell{\textbf{\# G.T. (M)}} &
    \textbf{Acc. (\%)} &
    \makecell{\textbf{\# G.T. (M)}} &
    \textbf{Acc. (\%)} &
    \makecell{\textbf{\# G.T. (M)}} \\
    \midrule

    \rowcolor{gray!36}
    \multicolumn{10}{@{}l}{
        \textbf{Non-Evolving Baselines}
    } \\

    \multirow{2}{*}{
        \makecell{Isolated Agents\\(ReAct)}
    }
    & --
    & 2.50 (0.29) & 34 (6)
    & 43.33 (0.00) & 0.5 (0.1)
    & 53.54 (3.11) & 2.6 (1)
    & 32.32 (2.14) & 17 (2) \\

    & + RAG
    & 5.26 (0.38) & 24 (3)
    & 32.22 (1.57) & 0.6 (0.2)
    & 48.32 (2.12) & 2.7 (2)
    & 30.90 (3.37) & \textbf{12 (3)} \\

    \midrule
    \rowcolor{gray!36}
    \multicolumn{10}{@{}l}{
        \textbf{Sequential Evolution Methods}
    } \\

    \multirow{3}{*}{
        \makecell[l]{Metacognitive Reuse}
    }
    & --
    & 4.09 (0.16) & 32 (5)
    & 46.67 (0.00) & 0.5 (0.1)
    & 57.74 (3.12) & 3.1 (1)
    & 33.05 (3.23) & 15 (2) \\

    & + Pooling
    & 4.11 (0.13) & 30 (4)
    & 46.67 (0.00) & 0.4 (0.1)
    & 60.10 (1.24) & 2.9 (1)
    & 36.33 (2.32) & 15 (2) \\

    & + RAG
    & 4.17 (0.12) & 30 (4)
    & 48.89 (1.57) & 0.5 (0.2)
    & 59.43 (1.26) & 2.7 (1)
    & 34.22 (4.32) & 15 (3) \\

    \multirow{3}{*}{HyperAgents}
    & --
    & 6.13 (0.23) & 28 (3)
    & 50.00 (0.00) & 0.6 (0.2)
    & 60.44 (1.33) & 2.5 (2)
    & 34.57 (3.13) & 14 (3) \\

    & + Pooling
    & 6.21 (0.56) & 27 (4)
    & 51.11 (1.57) & 0.5 (0.1)
    & 60.44 (1.19) & 2.4 (1)
    & 36.68 (2.13) & 15 (2) \\

    & + RAG
    & 6.27 (0.41) & 31 (3)
    & 52.22 (1.57) & 0.5 (0.2)
    & 60.10 (1.24) & 2.6 (1)
    & 35.45 (3.13) & 14 (2) \\

    \cdashline{1-10}

    ExpeL
    & --
    & 4.00 (0.32) & \textbf{21 (2)}
    & 43.33 (0.00) & 0.6 (0.1)
    & 52.36 (1.33) & 3.3 (1)
    & 35.26 (1.23) & 14 (3) \\

    AWM (online)
    & --
    & 2.42 (0.13) & 36 (4)
    & 36.67 (2.72) & 0.5 (0.1)
    & 49.49 (1.24) & 2.3 (1)
    & 34.69 (3.13) & 15 (2) \\

    BoT
    & --
    & 2.31 (0.23) & 42 (3)
    & 33.33 (0.00) & \textbf{0.3 (0.2)}
    & 50.17 (2.12) & 3.1 (1)
    & 29.48 (3.23) & 13 (2) \\

    \midrule
    \rowcolor{gray!36}
    \multicolumn{10}{@{}l}{
        \textbf{Federated Evolution Methods}
    } \\

    CORAL
    & --
    & 3.42 (0.23) & 45 (3)
    & 33.33 (0.00) & 0.8 (0.2)
    & 35.35 (1.24) & 4.3 (3)
    & 31.12 (1.12) & 18 (2) \\

    FederatedSkills
    & --
    & 1.32 (0.12) & 41 (5)
    & 36.67 (0.00) & 0.6 (0.2)
    & 49.33 (3.12) & \textbf{1.1 (0.9)}
    & 29.32 (2.34) & 20 (2) \\

    \midrule
    \textbf{Local} & \textbf{Server} & & & & & & & & \\
    \midrule

    \rowcolor{gray!36}
    \multicolumn{10}{@{}l}{
        \textbf{Various Local-Reasoning Strategies Plugged into FoT}
    } \\

    Metacognitive Reuse
    & \multirow{4}{*}{FoT Aggregation}
    & 8.41 (0.53) & 31 (3)
    & 43.33 (0.00) & 0.4 (0.2)
    & 64.14 (1.43) & 3.1 (1)
    & 41.33 (1.23) & 14 (3) \\

    HyperAgents
    &
    & \textbf{10.25 (1.32)} & 33 (4)
    & 55.56 (1.57) & 0.4 (0.1)
    & 58.08 (1.24) & 2.8 (0.8)
    & \textbf{43.82 (3.12)} & 14 (2) \\

    ExpeL
    &
    & 6.55 (0.43) & 22 (3)
    & 54.44 (1.57) & 0.5 (0.2)
    & 56.57 (2.30) & 3.1 (2)
    & 42.65 (1.23) & 16 (2) \\

    AWM
    &
    & 4.86 (1.32) & 37 (5)
    & 55.56 (1.57) & 0.6 (0.1)
    & 57.58 (3.11) & 2.4 (1)
    & 42.81 (3.13) & 14 (1) \\

    \midrule
    \rowcolor{gray!36}
    \multicolumn{10}{@{}l}{
        \textbf{FoT with Various Global Aggregation Strategies}
    } \\

    \multirow{7}{*}{FoT Local}
    & FoT Aggregation on each agent
    & 3.62 (1.31) & 32 (5)
    & 43.33 (0.00) & 0.4 (0.2)
    & 57.74 (1.19) & 2.9 (0.9)
    & 34.79 (3.13) & 16 (2) \\

    \cdashline{2-10}

    & Pooling
    & 4.24 (0.32) & 27 (4)
    & 46.67 (0.00) & 0.4 (0.1)
    & 60.10 (1.09) & 3.1 (1)
    & 35.48 (2.13) & 17 (3) \\

    & RAG
    & 4.07 (0.56) & 34 (3)
    & 46.67 (0.00) & \textbf{0.3 (0.1)}
    & 60.27 (1.26) & 2.9 (1)
    & 36.59 (3.53) & 14 (3) \\

    \cdashline{2-10}

    & Compaction
    & 7.12 (0.43) & 32 (3)
    & 53.33 (0.00) & 0.4 (0.2)
    & 66.16 (3.11) & 2.5 (1)
    & 41.42 (2.38) & 13 (2) \\

    & CoD
    & 5.32 (1.43) & 31 (3)
    & 53.33 (0.00) & 0.5 (0.2)
    & 54.55 (1.24) & 3.3 (1)
    & 40.13 (1.32) & 15 (2) \\

    \cdashline{2-10}

    & FoT Aggregation (w/o \Cref{prompt:relation})
    & 6.98 (0.76) & 29 (3)
    & 53.33 (2.72) & 0.4 (0.1)
    & 65.32 (1.04) & 3.0 (1)
    & 40.38 (2.34) & 16 (1) \\

    & FoT Aggregation
    & 9.88 (0.96) & 33 (3)
    & \textbf{56.67 (2.72)} & \textbf{0.3 (0.1)}
    & \textbf{69.02 (0.95)} & 2.8 (1)
    & 43.32 (3.43) & 15 (2) \\

    \bottomrule
\end{tabular}
\end{adjustbox}
\end{table*}

\begin{wrapfigure}{r}{0.42\textwidth}
    \centering
    \vspace{-1em}
    \centering
    \includegraphics[width=\linewidth]{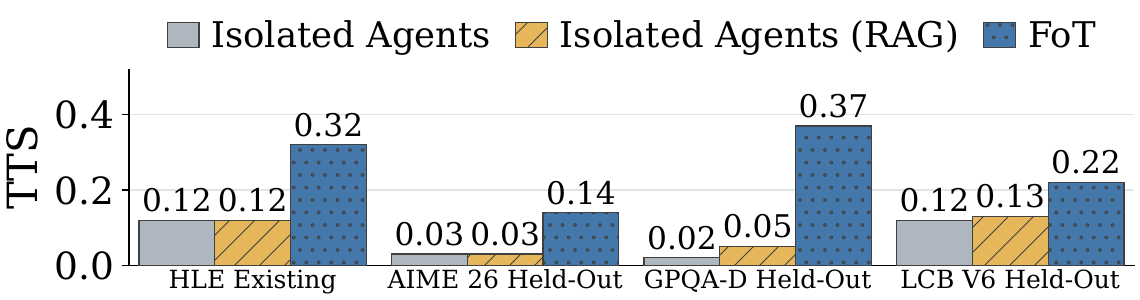}
    \captionsetup{skip=2pt}
    \caption{\small FoT achieves much higher TTS.}
    \label{tab:tts_multidomain}
    \vspace{-1.2em}
\end{wrapfigure}
\paragraph{True-Thinking Abilities.}
To examine whether FoT reduces decorative reasoning, we report the `true-thinking score'~\citep{zhao2025can} (TTS) in the multi-domain collaboration setting. TTS measures the proportion of reasoning steps whose removal would change the agent's answer, serving as a proxy for effective reasoning. Although storing all raw trajectories in a RAG datastore improves HLE accuracy, it does not improve TTS, suggesting that the RAG gain comes from retrieval rather than improved reasoning capability. FoT improves TTS, indicating that its insights help agents avoid redundant reasoning. 

\subsection{Research Insight Discovery}
\label{sec:insight}
We further apply FoT to open research question discovery, where agents collectively extract and synthesize insights across diverse areas of machine learning. Unlike end-to-end scientific agents that execute the full research pipeline and generate submission-ready papers~\citep{lu2024ai,schmidgall2025agent,jansen2025codescientist,weng2025deepscientist,yang2026aris,lyu2026evoscientist,song2026drclaw}, we focus only on discovering research insights. Each agent reads one paper and identifies research questions, after which FoT curates an insight library of generalizable ideas. We evaluate whether these insights could guide future papers as core contributions using three criteria: concrete methodological usage, how the method is presented, and whether the core contribution would fundamentally change without the insight; see \Cref{prompt:criteria}. Methods requiring explicit labels are not applicable because paper reading provides no explicit success signal.

We use Gemini 2.5 Flash\footnote{Gemini 2.5 Flash and Gemini 3.1 Pro have cutoff dates before the ICLR~2026 submission date.} to read ICLR~2025 papers and evaluate on ICLR~2026 papers. Baselines used for daily tasks are unsuitable because they are designed for agent skills or task solving and tend to produce generic paper-reading skills or one templated skill per paper. We therefore compare against three baselines: (1) standalone Gemini generates the top 200 insights from ICLR keywords under the same insight definition; (2) RAG~\citep{lewis2020retrieval} retrieves from all selected prior-year papers, using the same token budget as the reasoning traces, and generates 200 insights; and (3) FoT with its insight library explicitly limited to 200. Results are reported in \Cref{tab:iclrpaper} and cross-validated using Gemini 3.1 Pro (High), GPT-5.6 Sol (High), and Claude Opus 5 (High). We additionally conduct a blinded human evaluation. We select 50 accepted ICLR~2026 papers with the highest overall assessment scores and ask the evaluator to determine, using the same criteria, whether each insight library could guide each paper. The human results show strong agreement with the LLM judges, while FoT achieves the highest detection rate on the sampled papers. 

\begin{table}[t]
    \centering
    \caption{Percentage of papers guided by insights under different judge models. 
    Agreement is measured as the percentage of samples marked consistently as true or false by both judges.}
    \vspace{0.5em}
    \label{tab:iclrpaper}
    \scriptsize
    \setlength{\tabcolsep}{5pt}

    \begin{adjustbox}{max width=\linewidth}
        \begin{tabular}{@{}lcccccccccc@{}}
            \toprule
            \textbf{Method}
            & \makecell{\textbf{Gemini}\\\textbf{3.1 Pro} (\%)}
            & \makecell{\textbf{GPT-5.6}\\\textbf{Sol} (\%)}
            & \makecell{\textbf{Opus 5}\\(\%)}
            & \makecell{\textbf{Human}\\(\%)}
            & \makecell{\textbf{Gemini-}\\\textbf{GPT} (\%)}
            & \makecell{\textbf{Gemini-}\\\textbf{Opus} (\%)}
            & \makecell{\textbf{GPT-}\\\textbf{Opus} (\%)}
            & \makecell{\textbf{Human-}\\\textbf{Opus} (\%)}
            & \makecell{\textbf{Human-}\\\textbf{Gemini} (\%)}
            & \makecell{\textbf{Human-}\\\textbf{GPT} (\%)} \\
            \midrule

            Direct Generation
            & 58.49 & 86.97 & 50.40 & 40.00
            & 67.00 & 67.69 & 88.79
            & 74.00 & 68.00 & 82.00 \\

            RAG (Papers)
            & 74.49 & 80.68 & 86.69 & 44.00
            & 71.00 & 70.30 & 82.51
            & 76.00 & 70.00 & 84.00 \\

            RAG (Summaries)
            & 76.68 & 84.32 & 88.00 & 44.00
            & 73.37 & 73.33 & 86.99
            & 82.00 & 72.00 & 86.00 \\

            FoT
            & \textbf{94.58} & \textbf{99.53} & \textbf{93.50} & \textbf{56.00}
            & 94.47 & 88.99 & 93.29
            & 78.00 & 74.00 & 88.00 \\

            \bottomrule
        \end{tabular}
    \end{adjustbox}
\end{table}

\section{Properties of FoT}
\label{sec:properties}
In this section, we highlight other intriguing properties of FoT, including transferability to new models (\Cref{sec:transfer}), safeguarding raw task instances (\Cref{sec:privacyintegrity}), and personalization (\Cref{sec:personalization}).

\subsection{Cross-Model Transferability and Self-Correction}
\label{sec:transfer}

In our previous experiments, we have demonstrated that the library curated in one application can be transferred to held-out tasks or even new domains. Here, in \Cref{tab:transfer}, we further demonstrate that an insight library constructed by one LLM can be used to assist agents running another LLM.
This is practical when users upgrade to a stronger model or switch to a cheaper one while reusing the existing library. Notably, a library generated by a weaker model (Gemini 2.5 Flash Lite) can supervise and improve the performance of a stronger one (DeepSeek V4 Flash) (right), suggesting that a form of ``weak-to-strong generalization'' can occur through text-space knowledge transfer, beyond the gradient-space setting studied previously~\citep[e.g.,][]{burns2023weak}.

\begin{table*}[!htbp]
\centering
\small
\setlength{\tabcolsep}{3.8pt}
\renewcommand{\arraystretch}{1.08}
\captionsetup{skip=6pt}
\caption{
Cross-model transfer of shared knowledge. We compare directly pooled reasoning traces (`Pooling') with the insight library generated by FoT (`FoT').
Each arrow denotes the model used to construct the shared knowledge $\rightarrow$ the model used for downstream reasoning.
}
\label{tab:transfer}
\resizebox{\linewidth}{!}{
\begin{tabular}{@{}lccc@{\qquad}lccc@{}}
\toprule

\multicolumn{4}{c}{
\textit{DeepSeek V4 Flash $\rightarrow$ Gemini 2.5 Flash Lite}
}
&
\multicolumn{4}{c}{
\textit{Gemini 2.5 Flash Lite $\rightarrow$ DeepSeek V4 Flash}
}
\\

\cmidrule(lr){1-4}
\cmidrule(lr){5-8}

\makecell{\textbf{Transferred}\\\textbf{Insights}}
& \makecell{\textbf{PinchBench}\\\textbf{V1}}
& \makecell{\textbf{PinchBench}\\\textbf{V2 $\setminus$ V1}}
& \textbf{ClawEval}
&
\makecell{\textbf{Transferred}\\\textbf{Insights}}
& \makecell{\textbf{PinchBench}\\\textbf{V1}}
& \makecell{\textbf{PinchBench}\\\textbf{V2 $\setminus$ V1}}
& \textbf{ClawEval}
\\

\midrule

None (isolated)
& 34.78 (3.55)
& 33.33 (0.76)
& 19.25 (0.51)
&
None (isolated)
& 84.06 (5.42)
& 84.68 (1.32)
& 56.52 (0.51)
\\

Pooling
& 47.83 (3.55)
& 34.95 (0.76)
& 21.95 (0.59)
&
Pooling
& 73.91 (3.55)
& 69.35 (0.66)
& 43.89 (0.59)
\\

FoT
& \textbf{50.72 (4.10)}
& \textbf{54.30 (0.76)}
& \textbf{32.92 (0.51)}
&
FoT
& \textbf{89.86 (4.10)}
& \textbf{87.37 (0.76)}
& \textbf{69.36 (0.59)}
\\

\bottomrule
\end{tabular}
}
\vspace{-.5em}
\end{table*}

Interestingly, we also observe that Gemini 2.5 Flash Lite benefits from DeepSeek's pooled traces on PinchBench V1 (comparing `None' with `Pooling' on the left), whereas DeepSeek performs worse with Gemini's pooled traces (comparing `None' with `Pooling' on the right); in contrast, FoT aggregation improves over isolated reasoning in both cases. A likely reason is that the weaker Gemini model produces some erroneous reasoning traces, which are directly propagated by pooling, whereas FoT instead jointly reasons over multiple traces and synthesizes them into higher-level insights. We have the same observation in the multi-domain collaboration setting in Section~\ref{sec:mix3analysis}: although most traces produced by Gemini 2.5 Flash Lite originate from incorrect answers, their aggregated FoT library still improves downstream performance. 
We provide an example of such a self-correction process in \Cref{sec:supplytransferbility}. These results show that FoT can extract useful shared insights even from noisy local reasoning without external validation signals.

\subsection{Limiting Exposure of Raw Problem Instances}
\label{sec:privacyintegrity}
In addition to improving reasoning effectiveness and efficiency, FoT offers the side benefit of reducing exposure of raw problem instances.
\textit{First}, the uploaded abstracted reasoning traces do not contain the original problem or task descriptions. 
We use a strong multi-turn prompt-stealing attacker~\citep{sha2024prompt} to reconstruct agent inputs from reasoning traces. Token-level F1 scores between reconstructed and original inputs are below 0.25, and no personally identifiable information (PII) is leaked.
\textit{Second}, we verify that the insight library does not contain original question answers, preventing shortcut retrieval. Using strict 4-gram shingle matching, we observe near-{zero} Jaccard similarity between the insight library and ground-truth answers, and between reasoning traces and input questions. Full details and results are provided in \Cref{apx:privacy}. {While FoT is compatible with inference-time LLM differential privacy (DP) methods such as DP-Fusion~\citep{thareja2026dp} and InferDPT~\citep{tong2025inferdpt}, similar to how federated learning can be combined with DP protocols, we leave formal privacy guarantees to future work, as this is outside the scope of our work.}

\subsection{Personalization}
\label{sec:personalization}

FoT naturally supports personalization across agents. Although all agents receive the same shared insight library, each agent is prompted to identify and use only the insights relevant to its local task. As a result, different agents can automatically select different subsets and different numbers of insights from the same library, without separate agent-specific memories.

\vspace{-0.5em}
\begin{figure*}[!htbp]
    \centering
    \begin{minipage}[t]{\textwidth}
        \vspace{0pt}
        \centering
        \includegraphics[width=0.98\linewidth]{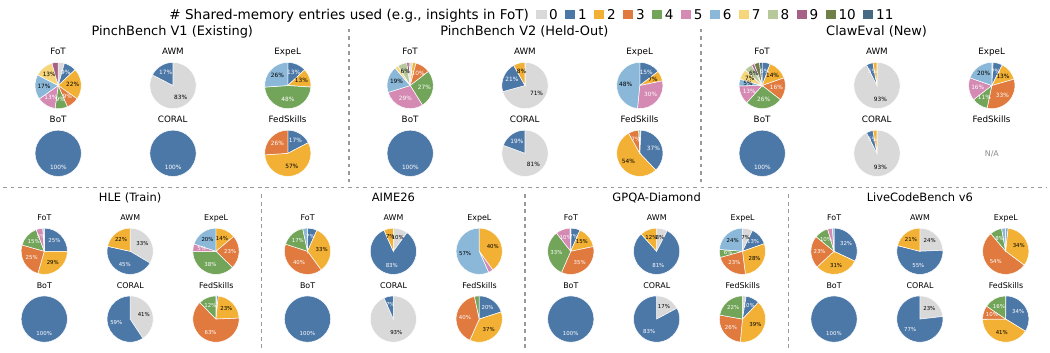}
        \captionsetup{skip=2pt}
        \vspace{0.5em}
        \captionof{figure}{The distribution of agents using different numbers of shared-memory entries for each method. For FoT, the shared memory is the insight library. FoT shows more flexible, task-dependent usage, with different agents selecting different numbers of relevant insights. 
        }
    \label{fig:distributioninsight}
    \end{minipage}
    \vspace{-0.5em}
\end{figure*}

We examine this behavior on the applications studied in Section~\ref{sec:applications} by recording the number of shared-memory entries used by each agent. For FoT, the shared memory is the insight library. We visualize the distributions of agents using different numbers of entries in the shared memory in \Cref{fig:distributioninsight}. It shows that different FoT agents select different numbers of insights depending on their tasks, illustrating task-dependent personalization from the same shared library.  In comparison, for AWM, most agents do not use any shared memory on held-out tasks, while other baselines typically use only a small fixed number of entries.  This is because the shared memory constructed by those methods is either too fine-grained (AWM, where one trace forms one insight) or too coarse-grained (CORAL, with one skill per major category), making selection on new tasks more challenging. FederatedSkills also considers personalization, but requires pre-defined task categories and category-specific execution. 
In \Cref{appx:personalization}, we further visualize which insights are used by each task in FoT, demonstrating diverse insight usage patterns. Overall, the results show that FoT provides personalization through automatic insight selection.

\section{Ablation Studies} \label{sec:ablation}

In this section, we empirically study the effects of several FoT parameters: the size of the insight library (\Cref{sec:librarysize}), agent participation rate (\Cref{sec:agentpct}), and local reasoning steps (Section~\ref{sec:ablation:local}). We use the application in \Cref{sec:openclaw}, unless otherwise specified.

\vspace{-0.5em}
\subsection{Effects of Insight Library Size}
\label{sec:librarysize}
\vspace{-0.5em}

\begin{figure*}[!htbp]
    \begin{minipage}[t]{0.485\textwidth}
        \vspace{0pt}
        \centering
        \includegraphics[width=0.98\linewidth]{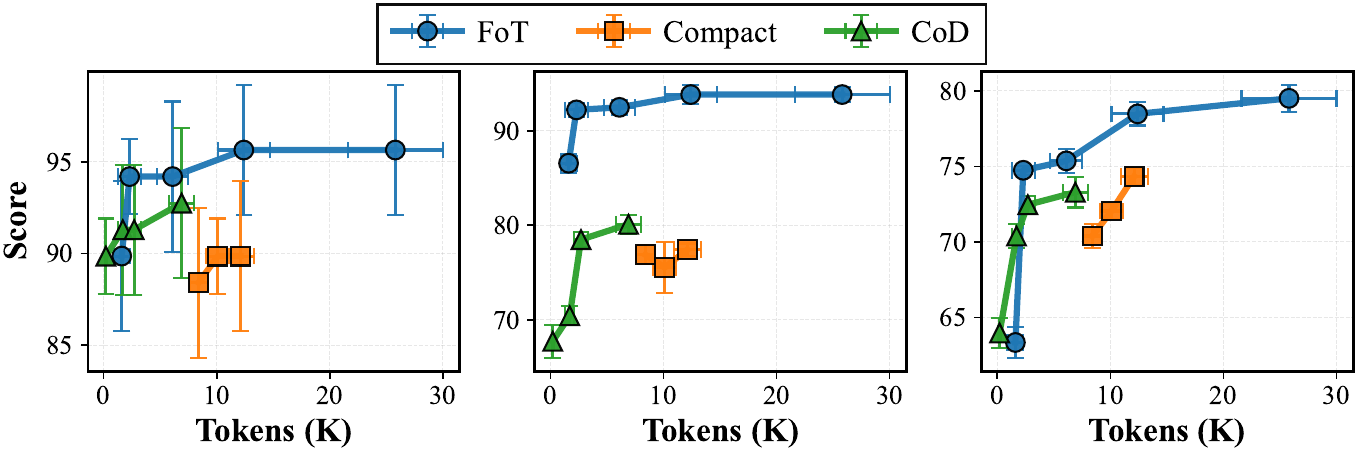}
        
        \captionsetup{skip=4pt}
        \vspace{0.5em}
        \captionof{figure}{The score/token tradeoff on existing, held-out, and new tasks in Section~\ref{sec:openclaw}. FoT achieves the best tradeoff and allows flexible control of the library size. 
        }
    \label{fig:numberofinsights}
    \end{minipage}
    \hfill
    \begin{minipage}[t]{0.485\textwidth}
        \vspace{0pt}
        \centering
        \includegraphics[width=0.98\linewidth]{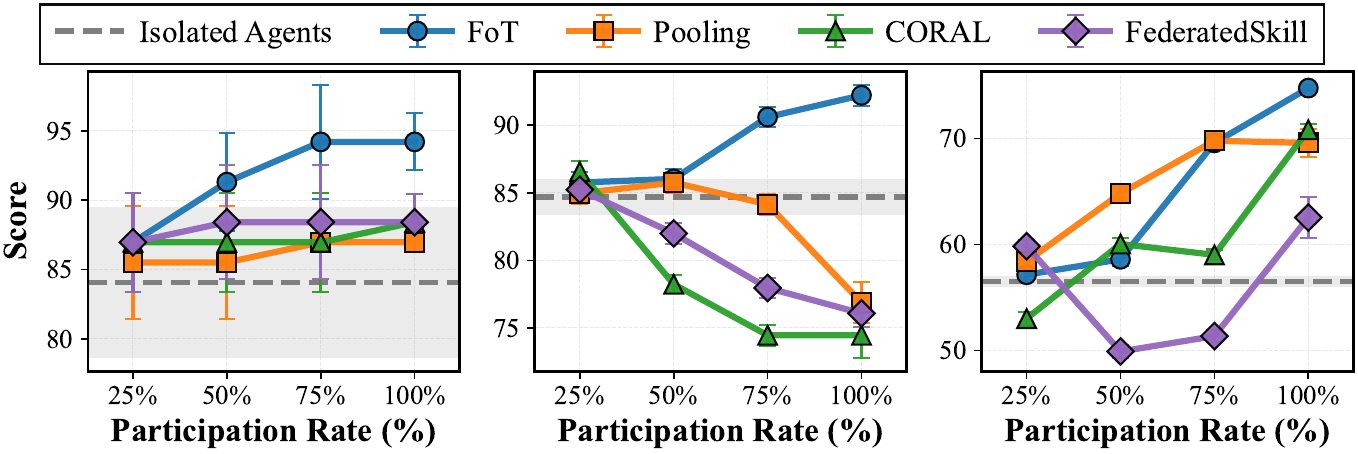}
        \captionsetup{skip=4pt}
        \vspace{0.5em}
        \captionof{figure}{Effects of agent participation rate on existing, held-out, and new tasks for different methods. CORAL and FederatedSkills can overfit to existing tasks.}
    \label{fig:pct}
    \end{minipage}
\end{figure*}

By default, \Cref{prompt:insight} allows the server to determine an appropriate number of insights automatically, producing libraries of 24.6 insights on average in our previous experiments. Here, we explicitly control the number of insights, resulting in libraries of different token sizes. We compare with other server-side context compression methods (Compaction and CoD) with varying levels of compression. 
In \Cref{fig:numberofinsights}, we observe that FoT achieves the best token/score tradeoff among all the methods. Further, FoT naturally reaches a decent performance score at $\sim$20 insights, as performance improves only marginally when increasing the library size from 20 to 200. {Such a result is expected: as shown in \Cref{sec:personalization}, most agents need around 2 insights, so 25 insights give sufficient coverage over the different tasks in PinchBench and ClawEval. Similarly, multi-domain collaboration involves over 3,000 tasks in total, and we observe that 85 insights are enough. Users can explicitly specify the number of insights in \Cref{prompt:insight} given the task statistics and distributions.}

\subsection{Effects of Agent Participation}
\label{sec:agentpct}
{In cross-device federated learning, only a subset of clients may participate in each aggregation round.} We further analyze the impact of varying the number of participating agents in FoT. As shown in \Cref{fig:pct}, higher participation rates lead to better performance. For the baselines, we observe possible over-fitting to existing tasks, as learned skills may not generalize well to tasks with different distributions, even demonstrating negative affects on reasoning (middle). {We also find that no single agent dominates the insight library, since performance consistently improves as participation increases.}

\subsection{Effects of Local Reasoning Rounds}
\label{sec:ablation:local}
\begin{figure}[!htbp]
    \vspace{-1em}
    \centering
    \includegraphics[width=\linewidth]{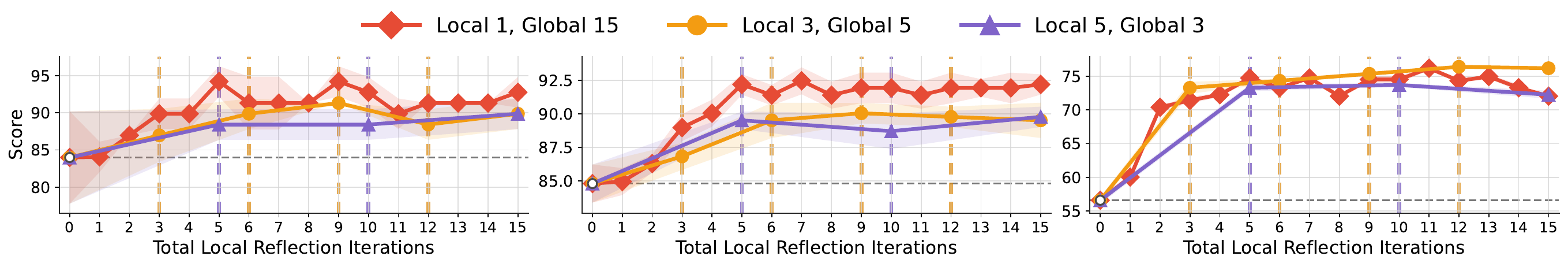}
    \captionsetup{skip=2pt}
    \vspace{-0.5em}
    \caption{The score on existing, held-out, and new tasks when client agents conduct different numbers of local reasoning iterations before the server aggregates. Less frequent server aggregation slightly lowers the performance. All settings converge after about 5 total local iterations.}
    \label{fig:round}
\end{figure}

{Motivated by the analogy to federated learning (FL), we propose to study a variant in which client agents perform multiple rounds of local self-evolution before uploading only the final-round reasoning traces for server aggregation. To study aggregation frequency, we fix the budget at 15 local reasoning rounds and aggregate every 1, 3, or 5 rounds. As shown in \Cref{fig:round}, less frequent aggregation leads to a slight performance drop, similar to the trends in FL. However, it reduces trace communication costs and the number of server-side aggregation operations. This result further shows that the conceptual analogy between FoT and FL can inspire algorithms with different trade-offs between final performance and communication/aggregation costs.}

\section{Conclusion, Limitations, and Future Work}
\label{sec:conclusion}
In this work, we proposed FoT, a federated-like multi-agent reasoning framework in which agents iteratively share and aggregate local reasoning traces to build a reusable insight library. We demonstrated the effectiveness of FoT across real-world daily tasks, cross-domain collaboration, and research insight discovery. Our analyses further show how aggregated insights improve reasoning and how FoT can accommodate different local reasoning strategies and server-side aggregation methods.

Our current experiments also suggest that FoT can limit direct exposure of raw problem instances. However, we do not study broader privacy risks or provide formal privacy guarantees. Since communication in FoT occurs in plaintext, clients can inspect and control what information is shared; developing stronger privacy-preserving mechanisms on top of this interface is an important direction for future work. Future work could also explore automating the design of FoT's reasoning and aggregation harnesses. 



\bibliography{main}
\bibliographystyle{unsrtnat}

\appendix
\crefalias{section}{appendix}

\newpage
\etocdepthtag.toc{mtappendix}
\etocsettagdepth{mtchapter}{none}
\etocsettagdepth{mtappendix}{subsection}
{
\parskip=0em
\tableofcontents
}


\newpage



\section{Pseudo Algorithms}
\label[appendix]{sec:alg}

We summarize the pseudocode of federation over (accumulated) gradients and federation over text in Algorithms~\ref{alg:fedavg} and~\ref{alg:fot}. 

\begin{figure}[!htbp]
	\begin{minipage}[t]{0.48\linewidth}
		\begin{algorithm}[H]
			\caption{Federation over Gradients}
            \label{alg:fedavg}
			\textbf{Input}: Initial global model, client datasets
            
			\For{each iteration}{
				\For{each \textbf{Client}}{
						Local optimization starting from the current global model \\
					Upload new model weights
				}
				\textbf{Server:}
				
				Collect all local model weights  \\
				Aggregation over model weights \\
				Broadcast the global model to clients
			}
			\textbf{Output:} Global model \\
		\end{algorithm}
	\end{minipage}
	\hfill
	\begin{minipage}[t]{0.49\linewidth}
		\begin{algorithm}[H]
			\caption{Federation over Text}
            \label{alg:fot}
			\textbf{Input}: Empty insight library, agent tasks, base LLM (can be different across agents)
			
			\For{each iteration}{
				\For{each \textbf{Agent}}{
						Local reasoning given the base LLM and current insight library \\
					Upload reasoning traces to the server
				}
				\textbf{Server:}
				
				Gather and aggregate all reasoning traces \\
				Update the insight library \\
				Broadcast the new library to agents
			}
			\textbf{Output:} Insight library\\
		\end{algorithm}
	\end{minipage}
\end{figure}

\section{Analogies Between Federated/Distributed Learning and FoT}
\label{sec:implication}

We note that there are rich connections between FL and FoT. In particular, in Table~\ref{tab:comparison}, we summarize how their algorithmic components can be mapped to each other. For instance, in FL, clients may adopt different optimization methods to solve local subproblems. Analogously, in FoT, each agent may leverage distinct local reasoning strategies and prompt designs to generate traces.

\begin{table}[h!]
\vspace{-.5em}
\centering
\caption{\small Connections between Federated Learning over Gradients and Federation over Text (FoT).}
\label{tab:comparison}
\small
\renewcommand{\arraystretch}{1.1}
\setlength{\aboverulesep}{0pt}
\setlength{\belowrulesep}{0pt}
\setlength{\extrarowheight}{0.3ex}
\begingroup
\rowcolors{2}{white}{gray!7}
\begin{tabularx}{\textwidth}{l XX }
\toprule
\textbf{Components} & \textbf{Federation over Gradients} & \textbf{Federation over Text (FoT)} \\
\midrule
Learning Objective & Learn a global model that generalizes to test data or new clients & Learn an insight library that improves the reasoning of agents \\
Learning Work & Local training via \textit{any} optimization method & Local task solving via \textit{any} self-improvement approach \\
Data Minimization & Share model updates rather than raw data & Share abstracted reasoning traces rather than raw instances \\
Local $\to$ Server & Clients upload model weights or gradients & Agents upload reasoning traces from problem-solving \\
Server Aggregation & Averaging over model weights or gradients & Aggregating over reasoning traces (text) using curated prompts \\
Server $\to$ Local & Current learned global model & Current curated insight library \\
Personalization & Maintaining personalized local models & Addressing diverse local tasks across different domains \\
\bottomrule
\end{tabularx}
\endgroup
\vspace{-1em}
\end{table}

\section{Complete Experiments for Applications of FoT}
\subsection{Real-World Daily Tasks}
\subsubsection{Cost Evaluation for the Application of OpenClaw}
\label[appendix]{sec:cost}

\begin{table*}[!h]
\centering
    \caption{
    Estimated computation and communication cost of all evaluated methods.
    Existing-task cost combines ordinary task execution with the additional
    client-side cost of generating reasoning traces and performing local
    self-evolution. Held-out- and new-task costs are task execution only. Server extra cost captures centralized updating, aggregation, compression, or datastore construction. Communication reports the token-equivalent payload uploaded to and
    downloaded from the server in each iteration.
    }
    \label{tab:all_method_costs}

 \footnotesize
\setlength{\tabcolsep}{2.4pt}
\renewcommand{\arraystretch}{1.12}

\begin{adjustbox}{max width=\textwidth}
\begin{tabular}{@{}llcccccc@{}}
    \toprule
    &
    &
    \multicolumn{3}{c}{\textbf{Task Cost (USD)}} &
    \multicolumn{2}{c}{\textbf{Communication}} \\
    \cmidrule(lr){3-5}
    \cmidrule(lr){6-7}

    \textbf{Method} &
    \textbf{Variant} &
    \makecell{\textbf{Extra Cost}\\\textbf{(USD)}} &
    \makecell{\textbf{Existing}\\\textbf{PinchBench V1}\\\textbf{+ Client Extra}} &
    \makecell{\textbf{Held-Out}\\\textbf{PinchBench V2 $\setminus$ V1}} &
    \makecell{\textbf{Upload}\\\textbf{(K)}} &
    \makecell{\textbf{Download}\\\textbf{(K)}} \\
    \midrule

    \rowcolor{gray!36}
    \multicolumn{7}{@{}l}{\textbf{Non-Evolving Baselines}} \\

    \multirow{3}{*}{\makecell{Isolated Agents\\(ReAct)}}
    & --
    & 0 & 0.046 & 0.916 & 0 & 0 \\

    & + Curated Skills
    & 0 & 0.078 & 0.939 & 0 & 6 \\

    & + RAG of raw traces
    & 0.001 & 0.060 & 0.968 & 54 (3) & 4 \\

    \midrule
    \rowcolor{gray!36}
    \multicolumn{7}{@{}l}{\textbf{Sequential Evolution Methods}} \\

    \multirow{3}{*}{
        \makecell[l]{Metacognitive Reuse~\citep{didolkar2025metacognitive}}
    }
    & --
    & 0 & 0.111 & -- & 0 & 0 \\

    & + Pooling
    & 0 & 0.104 & 0.885 & 39 (7) & 39 (7) \\

    & + RAG
    & 0.001 & 0.111 & 0.937 & 39 (7) & 4 \\

    \multirow{3}{*}{
        \makecell[l]{HyperAgents~\citep{zhang2026hyperagents}}
    }
    & --
    & 0 & 0.098 & -- & 0 & 0 \\

    & + Pooling
    & 0 & 0.093 & 1.023 & 22 (5) & 22 (5) \\

    & + RAG
    & $<0.001$ & 0.078 & 0.978 & 22 (5) & 4 \\

    \cdashline{1-7}

    ExpeL~\citep{zhao2024expel}
    & --
    & 0 & 0.088 & 0.820 & 21 (1) & 28 (3) \\

    AWM~\citep{wang2024agent}
    & --
    & 0 & 0.089 & 1.117 & 32 (9) & 19 (1) \\

    BoT~\citep{yang2024buffer}
    & --
    & 0 & 0.194 & 1.086 & 176 (10) & 118 (12) \\

    \midrule
    \rowcolor{gray!36}
    \multicolumn{7}{@{}l}{\textbf{Federated Evolution Methods}} \\

    CORAL~\citep{qu2026coral}
    & --
    & 0.132 & 0.072 & 0.932 & 50 (5) & 56 (3) \\
    FederatedSkills~\citep{yang2026federatedskill}
    & --
    & 0.264 & 0.068 & 0.894 & 45 (8) & 21 (2) \\

    \midrule
    \textbf{Local} & \textbf{Server} & & & & & \\
    \midrule

    \rowcolor{gray!36}
    \multicolumn{7}{@{}l}{
        \textbf{Various Local-Reasoning Strategies Plugged into FoT}
    } \\

    Metacognitive Reuse~\citep{didolkar2025metacognitive}
    & \multirow{4}{*}{FoT Aggregation}
    & \graycell{0.012}
    & \graycell{0.105}
    & \graycell{1.016}
    & \graycell{39 (7)}
    & \graycell{3 (1)} \\

    HyperAgents~\citep{zhang2026hyperagents}
    &
    & \graycell{0.009}
    & \graycell{0.086}
    & \graycell{1.108}
    & \graycell{22 (5)}
    & \graycell{4 (1)} \\

    ExpeL~\citep{zhao2024expel}
    &
    & \graycell{0.012}
    & \graycell{0.086}
    & \graycell{1.042}
    & \graycell{21 (1)}
    & \graycell{4 (1)} \\

    AWM~\citep{wang2024agent}
    &
    & \graycell{0.009}
    & \graycell{0.081}
    & \graycell{1.025}
    & \graycell{32 (9)}
    & \graycell{5 (1)} \\

    \midrule
    \rowcolor{gray!36}
    \multicolumn{7}{@{}l}{
        \textbf{FoT with Various Global Aggregation Strategies}
    } \\

    \multirow{7}{*}{FoT Local}
    & FoT Aggregation on each agent
    & 0 & 0.093 & 0.967 & 0 & 0 \\

    \cdashline{2-7}

    & Pooling
    & 0 & 0.089 & 0.888 & 28 (3) & 28 (3) \\

    & RAG
    & 0.001 & 0.075 & 0.907 & 28 (3) & 4 \\

    \cdashline{2-7}

    & Compaction~\citep{anthropic2024claudecookbooks}
    & 0.010 & 0.071 & 0.942 & 28 (3) & 10 (1) \\

    & CoD~\citep{adams2023sparse}
    & 0.008 & 0.069 & 0.855 & 28 (3) & 0.2 \\

    \cdashline{2-7}

    & FoT Aggregation (w/o \Cref{prompt:relation})
    & 0.013 & 0.078 & 1.051 & 28 (3) & 2 (1) \\

    & \graycell{FoT Aggregation}
    & \graycell{0.013}
    & \graycell{0.078}
    & \graycell{1.051}
    & \graycell{28 (3)}
    & \graycell{2 (1)} \\

    \bottomrule
\end{tabular}
\end{adjustbox}
\end{table*}
As agents relying on API services incur monetary costs%
\footnote{We follow OpenRouter's pay-as-you-go pricing: \url{https://openrouter.ai/pricing}.},
cost efficiency is a critical concern. We therefore examine whether the additional steps of self-evolution, including client-side reflection, reasoning-trace generation, and server-side insight-library aggregation, introduce substantial overhead. For existing tasks, client-side cost includes both task completion and reasoning-trace generation. For held-out and new tasks, the additional cost arises only from inserting shared content into the input prompt (no reasoning-trace generation cost). LLM-as-a-judge cost is excluded as it is not part of the task execution.

Server-side costs vary by aggregation strategy. FoT incurs additional LLM costs for insight aggregation, whereas pooling requires no additional model calls and RAG incurs API costs for embedding generation. Because we use the open-source OpenClaw framework, we exclude framework costs. Beyond model usage, federated methods also incur communication costs from uploading reasoning traces and downloading shared content. From \Cref{tab:all_method_costs}, FoT barely increases the extra server-side cost and only marginally increases task costs, while greatly reducing the download cost.

\subsubsection{Examples of Tool Usage}
\label[appendix]{apx:toolcallexample}
\paragraph{Example 1: Tech Conference Research.}
The first example is Tech Conference Research from PinchBench. The client needs to find five upcoming technology conferences and create \texttt{events.md} with
the name, date, location, and official website of each conference. For isolated agents, the client agent repeatedly revisits candidate sources, exhausts its
execution time after 79 tool calls, and receives a score of 0.00. In
Iteration~5, the insight library encourages an explicit search-fallback policy
and a report-first structure. The client uses only 39 calls and receives a score of 0.98. We can also see that the first insight, Progressive Search Fallback, is not originally related to the topic of searching for tech conferences; nevertheless, the agent can effectively identify and use it to avoid repeated searching and complete the task on time.

\begin{minipage}{\linewidth}
\begin{tcolorbox}[title=\textbf{Potentially Related Insight},fonttitle=\bfseries\scriptsize,width=\linewidth, enlarge left by=0pt, enlarge right by=0pt]
[insight] \textbf{Progressive Search Fallback}: When web search repeatedly times out or
fails, stop retrying the same route and pivot to direct HTTP requests to known authoritative sources, such as official event websites. Extract structured facts from page metadata or visible text. When to use: Apply when the primary search path is unreliable or repeatedly returns incomplete results. Details: 1) Begin with a bounded broad-search pass. 2) Select the candidates needed for the deliverable. 3) Retrieve each candidate's authoritative page directly. 4) Extract only the fields required by the output schema.

\ 

[insight] \textbf{Structured Report Skeleton}: Before collecting a large set of facts,
translate the user's requirements into the complete output structure. When to use: Apply to research tasks that require an exact number of entries or a fixed collection of fields. Details: 1) Define the table columns or report
sections first. 2) Populate them sequentially. 3) Record an authoritative source for every entry. 4) Write the artifact before the tool budget is exhausted. 5) Read the completed artifact back and verify every constraint.
\end{tcolorbox}
\end{minipage}

\begin{minipage}{\linewidth}
\begin{tcolorbox}[title=\textbf{Isolated Agents},fonttitle=\bfseries\scriptsize,width=\linewidth, enlarge left by=0pt, enlarge right by=0pt]
The client continues researching after it already has multiple plausible
conference candidates. The final eight calls demonstrate the repeated and
unstructured retrieval patterns:

\noindent\textbf{Call 72.}~~function name: \textsc{web\_fetch}.\\
argument:
\begin{verbatim}
{ "url": "https://www.ces.tech/attend/register", "max_chars": 8000 }
\end{verbatim}

\noindent\textbf{Call 73.}~~function name: \textsc{web\_search}.\\
argument:
\begin{verbatim}
{ "query": "tech conferences late 2026 upcoming list November December" }
\end{verbatim}

\noindent\textbf{Calls 74--78.}~~function names:
\textsc{web\_fetch}.\\
arguments: pages for Microsoft Build, Dreamforce, Wikipedia's
\emph{Web Summit} article, the official Web Summit site, and then the same
Wikipedia article again. The repeated retrieval does not advance a fixed
five-row output plan.

\noindent\textbf{Call 79.}~~function name: \textsc{web\_search}.\\
argument:
\begin{verbatim}
{ "query": "Web Summit Lisbon Portugal location venue" }
\end{verbatim}
\end{tcolorbox}
\end{minipage}

\begin{minipage}{\linewidth}
\begin{tcolorbox}[title=\textbf{FoT (Round 5)},fonttitle=\bfseries\scriptsize,width=\linewidth, enlarge left by=0pt, enlarge right by=0pt]
The client first establishes the workspace and date, performs a small
broad-search pass, pivots to authoritative pages, and reserves the final two
calls for writing and verification.

\noindent\textbf{Call 1.}~~function name: \textsc{list\_files}.\\
argument: \texttt{\{ path: "." \}}.

\noindent\textbf{Call 2.}~~function name: \textsc{run\_shell}.\\
argument: \texttt{date}.\\
output: establishes the temporal reference for ``upcoming.''

\noindent\textbf{Calls 3--6.}~~function name: \textsc{web\_search}.\\
arguments:
\begin{verbatim}
{ "query": "upcoming tech conferences 2026 2027 dates locations" }
{ "query": "major tech conferences fall 2026" }
{ "query": "tech conferences 2026 list dates" }
{ "query": "AWS re:Invent 2026 dates location" }
\end{verbatim}

\noindent\textbf{Calls 8--12.}~~function name: \textsc{web\_fetch}.\\
arguments: the discovered conference index and focused pages for AWS
re:Invent, Web Summit, and other selected candidates. Unlike the unsuccessful
run, these calls populate a predefined row schema rather than expanding the
candidate set indefinitely.

\noindent\textbf{Calls 32--37.}~~function names:
\textsc{run\_shell} and \textsc{web\_fetch}.\\
arguments: direct HTTP retrieval and lightweight text extraction from the
official Google Cloud Next, CES, re:Invent, and Dreamforce pages after ordinary
search/fetch routes were incomplete.

\noindent\textbf{Call 38.}~~function name: \textsc{write\_file}.\\
argument:
\begin{verbatim}
{
  "path": "events.md",
  "content": "# Upcoming Tech Conferences\n\n
| # | Conference | Date | Location | Website |\n
|---|------------|------|----------|---------|\n
| 1 | Dreamforce 2026 | Sep 15--17, 2026 | San Francisco, USA | ... |\n
... five conferences total ..."
}
\end{verbatim}

\noindent\textbf{Call 39.}~~function name: \textsc{read\_file}.\\
argument: \texttt{\{ path: "events.md" \}}.\\
output: verifies that the five required rows and all four requested fields are
present.
\end{tcolorbox}
\end{minipage}

\paragraph{Example 2: Google Workspace Task Management.}
The second example is \emph{GWS Task Management}, a held-out task from ClawEval. The agent needs to inspect the existing
Google Tasks list and recent Gmail messages, convert actionable messages into
tasks, mark an already completed item accordingly, and write
\texttt{task\_summary.md}. In isolated agents, the agent produces a verbose summary
containing zero extracted action items and performs none of the requested task
updates, receiving a score of 0.22. With the insight library (generated in round 2), the agent is guided to
inventory the entire queue, read complete message bodies, correlate related items, perform the mutations, and verify both remote and local state. The later attempt receives a score of 1.00.

\begin{minipage}{\linewidth}
\begin{tcolorbox}[title=\textbf{Potentially Related Insight},fonttitle=\bfseries\scriptsize,width=\linewidth, enlarge left by=0pt, enlarge right by=0pt]
[insight] \textbf{Inventory Before Prioritization}: For a high-volume email or incident queue, first enumerate all items and read their complete contents before acting. When to use: Apply when a task asks the agent to review a queue, identify actionable items, and update another system. Details: 1) List all
source items in one bounded request. 2) Retrieve the full contents of each candidate. 3) Score urgency, criticality, and sender authority. 4) Inspect the destination system before creating or changing records.

\ 

[insight] \textbf{Cluster Related Items}: Treat messages that refer to the same project,
incident, or deadline as one logical unit for prioritization and action. When to use: Apply when several emails may describe the same underlying work. Details: 1) Identify common entities, events, and deadlines. 2) Separate
distinct requests from repeated notifications. 3) Create one well-scoped task for each distinct action instead of duplicating work.

\ 

[insight] \textbf{Verify Deliverable and State}: After changing an external system or
creating a file, retrieve the resulting state and compare it against every requested constraint. When to use: Apply whenever success depends on both a remote mutation and a local artifact. Details: 1) Re-list the remote records. 2) Confirm created and completed items. 3) Write the requested summary. 4) Read
the summary back before reporting completion.
\end{tcolorbox}
\end{minipage}

\begin{minipage}{\linewidth}
\begin{tcolorbox}[title=\textbf{Isolated Agents},fonttitle=\bfseries\scriptsize,width=\linewidth, enlarge left by=0pt, enlarge right by=0pt]
\noindent\textbf{Calls 1--44.}~~The client repeatedly explores command setup,
Gmail access, and Tasks access but never establishes a complete inventory of
the messages and existing tasks. Consequently, it cannot correlate messages
with concrete action items or identify the already completed task.

\noindent\textbf{Call 45.}~~function name: \textsc{write\_file}.\\
argument: a verbose \texttt{task\_summary.md} reporting zero actionable emails,
zero reviewed tasks, zero completed tasks, and zero newly created tasks.
\end{tcolorbox}
\end{minipage}

\begin{minipage}{\linewidth}
\begin{tcolorbox}[title=\textbf{FoT (Round 2)},fonttitle=\bfseries\scriptsize,width=\linewidth, enlarge left by=0pt, enlarge right by=0pt]
The later client follows a complete inventory--act--verify pipeline.

\noindent\textbf{Call 1.}~~function name: \textsc{run\_shell}.\\
argument:
\begin{verbatim}
gws tasks tasklists list
\end{verbatim}

\noindent\textbf{Call 2.}~~function name: \textsc{run\_shell}.\\
argument:
\begin{verbatim}
gws tasks tasks list --params '{"tasklist":"default"}'
\end{verbatim}
output: inventories the current ``My Tasks'' list before any mutation.

\noindent\textbf{Call 3.}~~function name: \textsc{run\_shell}.\\
argument:
\begin{verbatim}
gws gmail users messages list \
  --params '{"userId":"me","maxResults":50}'
\end{verbatim}
output: retrieves one bounded inventory instead of repeatedly searching for
individual guessed messages.

\noindent\textbf{Call 4.}~~function name: \textsc{run\_shell}.\\
argument: loop over the identified actionable message IDs and call
\texttt{gws gmail users messages get} with \texttt{format=full}.\\
output: complete bodies for the four actionable messages, enabling the client
to distinguish separate requests from messages about the same work item.

\noindent\textbf{Call 5.}~~function name: \textsc{run\_shell}.\\
argument:
\begin{verbatim}
gws tasks tasks patch \
  --params '{"tasklist":"default","task":"taskk002"}' \
  --json '{"status":"completed"}'
\end{verbatim}

\noindent\textbf{Call 6.}~~function name: \textsc{run\_shell}.\\
argument: one \texttt{gws tasks tasks insert} operation for each distinct email
action item, including \texttt{Investigate CI pipeline failure on main
(Build \#1234)}.

\noindent\textbf{Call 7.}~~function name: \textsc{run\_shell}.\\
argument:
\begin{verbatim}
gws tasks tasks list --params '{"tasklist":"default"}'
\end{verbatim}
output: verifies the completed existing task and the newly created tasks.

\noindent\textbf{Call 8.}~~function name: \textsc{write\_file}.\\
argument: \texttt{task\_summary.md}, organized into existing tasks, actionable
emails, completed work, and newly created tasks.

\noindent\textbf{Call 9.}~~function name: \textsc{run\_shell}.\\
argument: \texttt{cat task\_summary.md}.\\
output: verifies the local deliverable after the remote-state verification.
\end{tcolorbox}
\end{minipage}

\subsection{Multi-Domain Collaboration}
\label[appendix]{sec:supplymultidomain}
In the first example, we present an AIME26 problem for which Gemini 2.5 Flash Lite initially fails to produce the correct answer and becomes lost after several reasoning steps. The insight library contains three insights merged from multiple domains, including one derived specifically from an HLE problem. The first two insights come from computational linguistics and text-analysis clusters and provide high-level guidance on how an agent should approach such questions. The third insight is most closely related to computer science, but it transfers to this arrangement-and-counting problem by guiding the agent to identify the constraints step by step and enumerate all valid possibilities. We provide the complete raw reasoning process for each round in Appendix~\ref{appx:completereasoning}.

\begin{tcolorbox}[title=\textbf{Question: AIME26, Problem 20 (Answer: 190)},fonttitle=\bfseries\scriptsize]
An urn contains $n$ marbles. Each marble is either red or blue, and there are at least $7$ marbles of each color. When $7$ marbles are drawn randomly from the urn without replacement, the probability that exactly $4$ of them are red equals the probability that exactly $5$ of them are red. Find the sum of the five least values of $n$ for which this is possible. 
\end{tcolorbox}

\begin{minipage}{0.49\linewidth}
\begin{tcolorbox}[title=\textbf{Incorrect (Isolated Agents)},fonttitle=\bfseries\scriptsize,width=\linewidth, enlarge left by=0pt, enlarge right by=0pt]
The agent set up the hypergeometric equality $\binom{R}{4}\binom{B}{3}=\binom{R}{5}\binom{B}{2}$ and reduced it correctly to $5(B-2)=3(R-4)$, i.e. $3R-5B=2$, even finding the general solution $(R,B)=(5m+4,\,3m+2)$. It then lost the objective: it started asking for ``the ratio of red to blue marbles'', never re-introduced $n=R+B$, cycled through $(14,8),(19,11),(24,14)$ debating which ``ratio'' the problem intended, and terminated without a final answer.
\end{tcolorbox}
\end{minipage}
\hfill
\begin{minipage}{0.49\linewidth}
\begin{tcolorbox}[title=\textbf{Correct (FoT, Round 1)},fonttitle=\bfseries\scriptsize,width=\linewidth, enlarge left by=0pt, enlarge right by=0pt]
The agent first enumerated the three conditions and the exact target (``sum of the five least values of $n$''). It derived the same relation $5B+2=3R$, but immediately eliminated $R,B$ in favor of the target variable: $R=(5n+2)/8$, $B=(3n-2)/8$. Integrality gives $n\equiv 6 \pmod 8$; the constraints $R\ge 7$, $B\ge 7$ give $n\ge 19.3$; combining yields $n=22,30,38,46,54$ and the sum $190$.
\end{tcolorbox}
\end{minipage}

\begin{tcolorbox}[title=\textbf{Potentially Related Insights},fonttitle=\bfseries\scriptsize]
[Insight] \textbf{Defining Problem Scope And Constraints} (distilled from cluster: ``Computational Linguistics and Text Analysis''): Clearly define the problem by listing all filtering criteria (attributes, required values) derived from the query. Ensures all conditions are acknowledged for precise entity identification. Use this at the start of any problem requiring identification of a specific entity matching a profile.

\ 

[Insight] \textbf{Sequential Constraint Propagation} (distilled from cluster: ``Computational Linguistics and Text Analysis''): Solve problems with multiple criteria by applying constraints sequentially to filter potential solutions, followed by final verification. Use this for queries with layered requirements, especially when constraints can be applied in a natural order.

\ 

[Insight] \textbf{Consolidate Bounds And Handle Parameter Edge Cases} (distilled from cluster: ``Computational Complexity and Theoretical Computer Science''): Combine all derived inequalities for $k$ to establish the final VC dimension bound, refining for parameter edge cases and ensuring validity. Use this for formulating the final, universally applicable VC dimension, addressing degenerate scenarios.

\ 

[Insight] \textbf{Constraint Driven Pruning And Filtering} (distilled from HLE problem 1: chess mate-in-N puzzle): Systematically identify and eliminate candidate options that directly violate explicitly stated rules or constraints. This process reduces the search space by discarding invalid choices early, allowing focus on permissible and potentially viable ones. Use this at the outset of any problem with explicit restrictions on first moves, piece movements, or configuration rules to prune the search space effectively.
\end{tcolorbox}

In the second example, insights from one domain are applied to another domain. Both runs address the same chemistry problem. The isolated agent mistranscribes its own conclusion (toluene $\rightarrow$ benzene) and, instead of treating the mismatch with the option list as a signal to re-check, rationalizes it. FoT leverages three insights from number theory and mathematics, where a mathematical framing habit and a logician's condition-checking habit correct an organic-chemistry answer.

\begin{tcolorbox}[title=\textbf{Question: GPQA Diamond, Problem 197 (Answer: B)},fonttitle=\bfseries\scriptsize]
Compound X, which has the following IR and NMR data, reacts with red phosphorus and HI. Identify the final product.
IR: 3400--2500 cm$^{-1}$, 1720 cm$^{-1}$, 1610 cm$^{-1}$, 1450 cm$^{-1}$
$^1$H NMR: 10.5 ppm (bs, 1H), 8.0 ppm (d, 2H), 7.2 ppm (d, 2H), 2.9 ppm (m, 1H), 1.7 ppm (m, 2H), 1.4 ppm (d, 3H), 0.9 ppm (t, 3H).

Options: A. 2-(4-ethylphenyl)propanoic acid  B. 1-(sec-butyl)-4-methylbenzene  C. 4-(sec-butyl)benzoic acid  D. 1-isobutyl-4-methylbenzene
\end{tcolorbox}

\begin{minipage}{0.49\linewidth}
\begin{tcolorbox}[title=\textbf{Incorrect (Isolated Agents)},fonttitle=\bfseries\scriptsize]
The agent correctly read the spectra (carboxylic acid, para-disubstituted ring, CH$_3$--CH$_2$--CH(CH$_3$)-- fragment) and identified X as 4-(sec-butyl)benzoic acid, and correctly recalled that red P/HI reduces --COOH to --CH$_3$ (``converts benzoic acid to toluene''). When it wrote the product down, however, it dropped the newly formed methyl group and named it ``4-(sec-butyl)benzene''. Finding that string absent from the options, it declared the problem defective and, ``if we are forced to choose'', picked the carboxylic acid 2-(4-ethylphenyl)propanoic acid because it ``represents a significant structural change''.
\end{tcolorbox}
\end{minipage}
\hfill
\begin{minipage}{0.49\linewidth}
\begin{tcolorbox}[title=\textbf{Correct (FoT, Round 2)},fonttitle=\bfseries\scriptsize]
The agent deduced the same X and the same R--COOH $\rightarrow$ R--CH$_3$ reduction, but then wrote the product explicitly as 4-(sec-butyl)toluene and listed its equivalent IUPAC names (1-methyl-4-(sec-butyl)benzene = 1-(sec-butyl)-4-methylbenzene) \emph{before} checking each option in turn: the two acids are rejected (one is the starting material, the other has the wrong skeleton), the isobutyl isomer is rejected by the NMR (a doublet at 1.4 and a triplet at 0.9 require sec-butyl), leaving 1-(sec-butyl)-4-methylbenzene. The agent marked three library insights as used.
\end{tcolorbox}
\end{minipage}
\begin{tcolorbox}[title=\textbf{Potentially Related Insight},fonttitle=\bfseries\scriptsize]
[Insight] \textbf{Problem Framing And Goal Definition} (distilled from cluster: ``Number Theory''): Clearly define mathematical objects, structures, and the precise question. Use this at the start of elliptic curve problems or similar analyses to ensure objective precision.

\

[Insight] \textbf{Mechanism Principle Congruence Analysis} (distilled from cluster: ``impossibility theorem''): Systematically map the core operational mechanism or principle of a theory, algorithm, or viewpoint against the precise definitions of relevant conditions (often derived from theorems or logical constraints). This comparison aims to identify precisely how and why the subject either upholds or violates each condition, assessing the significance of any violation. Use this when specific features of a theory or algorithm need to be evaluated against abstract principles or logical rules.

\ 

[Insight] \textbf{Goal Oriented Answer Selection Prioritization} (distilled from cluster: ``impossibility theorem''): Select the most accurate or relevant answer by prioritizing the violation or interaction most central to the subject theory's stated goal or primary function. This technique distinguishes fundamental arguments from tangential ones when multiple potential interpretations or violations exist. Use this when analysis yields multiple plausible options and a judgment of primacy is required, typically in multiple-choice scenarios.

[Insight] \textbf{Exhaustive Candidate Evaluation} (distilled from cluster: ``activation-function derivatives''): Systematically apply analytical procedures to each candidate function to ensure all options are considered. Use this for problems with discrete sets of options requiring property verification.
\end{tcolorbox}

When examining incorrect reasoning processes (e.g.,~\Cref{reasonaime26p20incorrect,reasongpqadiamondp197incorrect} in the Appendix), an interesting phenomenon occurs: there are steps that are entirely irrelevant to the final answer. For example, after correctly deriving the relation $3R-5B=2$, \Cref{reasonaime26p20incorrect} spends the rest of the trace debating which ``ratio of red to blue marbles'' the problem intends, a quantity that is never asked for, and then restates the same conclusion in five near-identical ``Finalizing Marble Ratio'' blocks without ever producing the requested sum. Similarly, \Cref{reasongpqadiamondp197incorrect} first identifies compound X and the P/HI reduction correctly, and then devotes several paragraphs to speculating about ``an intricate rearrangement'' or ``an error in the problem statement'' before selecting an option it explicitly states it cannot justify; none of these paragraphs changes the answer. As a result, such steps can be viewed as `decorative thinking' steps. Instead, with FoT's insight library, agents do not have to exhibit such decorative thinking in \Cref{reasonaime26p20correct,reasongpqadiamondp197correct}: both traces open by enumerating the constraints and the target quantity, proceed through the constraints in order, and finish with an explicit check of the candidates against the options, using fewer output tokens than their isolated counterparts on the chemistry problem. This indicates that insights as a strong and effective prior can help LLM agents quickly start from knowledge of previous problems, reducing the chance of falling into unnecessary or illogical steps. The experiment measuring the true-thinking score~\citep{zhao2025can} further verifies this argument.

\section{Complete Experiments for Properties of FoT}

\subsection{Transferability and Self-Correction}
\label[appendix]{sec:supplytransferbility}

\paragraph{Example: Email Search and Summarization.}
The agent needs to search a collection of
emails and synthesize the relevant information. The isolated agent gets a score of 0. The reasoning trace contains only one tool-call reflection.

\begin{minipage}{\linewidth}
\begin{tcolorbox}[title=\textbf{Potentially Faulty Reasoning Trace},fonttitle=\bfseries\scriptsize,width=\linewidth, enlarge left by=0pt, enlarge right by=0pt]
This strategy filters a collection of files using keywords or patterns in their filenames. It is an initial pruning step for isolating documents related to a known subject or project. It is recommended when many files must be narrowed before deeper content analysis, particularly when filenames contain explicit identifiers; for example, select emails whose filenames contain ``Project Alpha.''
\end{tcolorbox}
\end{minipage}

An issue with this reflected reasoning trace is that it makes an environmental assumption that
filenames reliably encode relevance but does not provide a fallback for vague, misleading, or missing identifiers. If an agent directly uses this trace in subsequent work, it may have issues when selecting emails by filename and miss some important files.
\begin{minipage}{\linewidth}
\begin{tcolorbox}[title=\textbf{Potentially Related Insights from the Library},fonttitle=\bfseries\scriptsize,width=\linewidth, enlarge left by=0pt, enlarge right by=0pt]
[insight] After identifying candidate documents, inspect their contents for the concepts and keywords specified by the user. Extract each relevant statement together
with its surrounding context so that a keyword match is not mistaken for supporting evidence. Use filename filtering only to reduce the search space; it is not a substitute for reading the selected files.

\medskip
[insight]
\path{insight_aggregate_and_prioritize_chronological_information}:
When several documents describe the same evolving project, collect their dated
claims, order them chronologically, and reconcile earlier statements with newer updates. Give priority to the most recent status while preserving older events that explain how the project changed. The final answer should present one coherent current account rather than a disconnected list of matches.
\end{tcolorbox}
\end{minipage}

By merging content from other domains, the insight library automatically corrects the issue in the previous reasoning trace related to filename filtering. It provides more general rules for identifying candidate documents and a more promising way to gather supporting evidence. The second insight is also related to the trace, providing guidance on what an agent should do when several documents describe the same project. The two insights supply the missing middle of the workflow. As a result, if we adopted previous approaches of directly adding a reasoning trace to the skill book, the faulty guidance would be propagated, whereas with our merging approach, we obtain better insights to guide future agents.

\subsection{Safeguarding Raw Problem Instances}
\label[appendix]{apx:privacy}
\begin{table}[!htbp]
	\centering
    \caption{Prompt stealing attack results. The attacker uses Gemini-3.6-Flash with PinchBench reasoning traces in FoT to reconstruct the original input. The low character-level similarity, low token-level F1, and low parameter-extraction accuracy indicate that the attack fails to recover the original input, suggesting an effective defense. No personally identifiable information (PII) is leaked. The closest reconstruction achieved by the attacker is inferring that the instruction involves common file names such as \emph{``ai blog.txt''} and \emph{``humanized blog.txt''}.}
	\label{tab:privacyattack}
    \setlength{\tabcolsep}{6pt}
\renewcommand{\arraystretch}{1.1}
		\scalebox{0.86}{\begin{tabular}{lcccc}
			\toprule
			Method  &Char Similarity&Token F1 &PII leaked&Parameter Extractor Acc\\
			\midrule
            Raw Transcripts&65.34\%&67.82\%&26&86.96\%\\
            \midrule
            Metacognitive Reuse&12.64\%&17.83\%&0&8.70\%\\
            HyperAgents&9.26\%&8.72\%&0&4.35\%\\
            ExpeL&23.68\%&59.32\%&8&73.91\%\\
            AWM&44.58\%&49.38\%&5&82.61\%\\
            FoT&10.92\%&16.38\%&0&13.04\%\\
            \midrule
            CORAL&43.89\%&56.78\%&0&78.26\%\\
            FederatedSkills&11.68\%&20.34\%&11&17.39\%\\
            
			\bottomrule
		\end{tabular}}
\end{table}

First, we investigate whether the original task description can be inferred from uploaded reasoning traces. We adopt a strong prompt-stealing attack setting~\citep{sha2024prompt}, where the attacker or an honest-but-curious server aims to reconstruct the original task prompt from observed traces. In PinchBench daily tasks, the attacker is given a reasoning trace and must infer the agent's input, including the user query and auxiliary information (e.g., documents to be summarized). The attack consists of two components: a parameter extractor, which classifies the prompt type into direct user request, role-based (persona or instructional framing), or in-context (demonstrations or substantial context), and a prompt extractor, which reconstructs the prompt. The two components interact for three rounds to refine the output.

We evaluate reconstruction using character-level similarity, token-level F1, and personally identifiable information (PII) leakage. For the 23 PinchBench tasks, we manually inspect PII leakage and find no recoverable personally identifiable information. We also report the accuracy of prompt-type prediction, which is below random guessing. As shown in Table~\ref{tab:privacyattack}, both character-level similarity and token-level F1 remain low.  
At the same time, these metrics are not expected to be zero. Reasoning traces necessarily retain some task-relevant information to enable knowledge transfer across agents. 
Prior work \citep{sha2024prompt} suggests that successful prompt reconstruction typically requires similarity exceeding 60\%, which is not observed in our setting.

\begin{table}[!htbp]
\centering
\small
\caption{The measurement of overlap between the insight library (IL), reasoning traces (RT), and problems. Jaccard similarity is computed between 4-gram token sets constructed from problem statements, answers, reasoning traces, and insight libraries.}
\label{tab:privacy}

\setlength{\tabcolsep}{6pt}
\renewcommand{\arraystretch}{1.1}
\scalebox{0.85}{
\begin{tabular}{l l c c}
\toprule
\textbf{Application} 
& \textbf{Method}
& \makecell{\textbf{Jaccard}\\\textbf{(RT and problems)}}
& \makecell{\textbf{Jaccard}\\\textbf{(IL and problems \& answers)}} \\
\midrule

\multirow{5}{*}{HLE}
& FoT                 & 0.010 & 0.000 \\
& Metacognitive Reuse & 0.018    & 0.003    \\
& HyperAgents         & 0.009    & 0.000    \\
& ExpeL               & 0.136    & 0.013    \\
& AWM                 & 0.159    & 0.008    \\
& CORAL               & 0.498    & 0.347    \\
& FederatedSkills     & 0.014    & 0.003    \\

\midrule

\multirow{5}{*}{Daily Tasks}
& FoT                 & 0.006 & 0.000 \\
& Metacognitive Reuse & 0.008    & 0.000    \\
& HyperAgents         & 0.012    & 0.000    \\
& ExpeL               & 0.214    & 0.000    \\
& AWM                 & 0.146    & 0.000    \\
& CORAL               & 0.201    & 0.113    \\
& FederatedSkills     & 0.008    & 0.001    \\
\midrule
Insights from ICLR 2025
& FoT & 0.000 & 0.000 \\
\bottomrule
\end{tabular}}
\end{table}

{Second, we directly measure textual overlap between the shared artifacts and the original task data. 
For each application, we concatenate all problem instances into a document and partition the document into 4-gram token shingles. Problems refer to questions, tasks, or papers depending on the application.  We do the same for answers, insight libraries, and reasoning traces, separately. Each document is a set of all 4-gram token shingles. We then evaluate similarities between (a) reasoning traces and problems, and between (b) the insight library and problems + answers, by reporting the Jaccard similarities between the two sets of 4-gram tokens in the corresponding two documents.  Prior work shows that the choice of $n$ significantly affects similarity estimates, with smaller $n$ leading to inflated overlap due to common phrases and larger $n$ being overly sensitive to minor edits~\citep{puspaningrum2020detection}. We adopt 4-gram shingles as a balanced choice.}

In \Cref{tab:privacy}, {we measure the Jaccard similarity for different methods' reasoning traces and insight libraries when we plug them into the FoT framework across applications.} We observe that our method's Jaccard similarity is near zero, indicating that the reasoning traces have little overlap with raw user questions and answers. The zero similarity between the insight library and the problems further suggests that an adversary cannot obtain raw problem instances by examining the library.  Similarities between uploaded traces and problems on AWM, ExpeL, and CORAL are high because they need to share the raw reasoning trajectories. CORAL and AWM explicitly need to aggregate over raw trajectories, while ExpeL needs several successful raw trajectories.

We emphasize that these experiments do not establish a general privacy guarantee, as privacy is a broad notion. On one hand, FoT has similar benefits to vanilla federated learning in terms of data minimization without transmitting raw trajectories. In fact, if we narrow down the scope to not leaking sensitive entities implicit in the reasoning traces (e.g., the metric of Sensitive Entity Leakage Rate in FederatedSkills~\citep{yang2026federatedskill}), FoT can satisfy the constraint.  On the other hand, reasoning traces may still reveal higher-level semantic information, such as the topic or structure of a task; for example, a trace mentioning a file named ``ai blog.txt'' may reveal that the task concerns writing about AI. Instead, FoT provides a form of data reduction: raw task instances and trajectories need not be transmitted to the server. Stronger privacy guarantees are complementary to our framework. 
For example, inference-time differential privacy mechanisms such as DP-Fusion~\citep{thareja2026dp} or InferDPT~\citep{tong2025inferdpt} could be applied locally before traces are uploaded, analogous to adding differential privacy mechanisms to federated learning. A broader study of semantic leakage and privacy defenses for multi-agent reasoning is left for future work.

\subsection{Personalization}
\label[appendix]{appx:personalization}
Here, we further demonstrate, for the application in \Cref{sec:openclaw}, how each task leverages the insights from the library aggregated at iteration 5 (best run).

\begin{figure}[!h]
    \centering
    \includegraphics[width=0.87\linewidth]{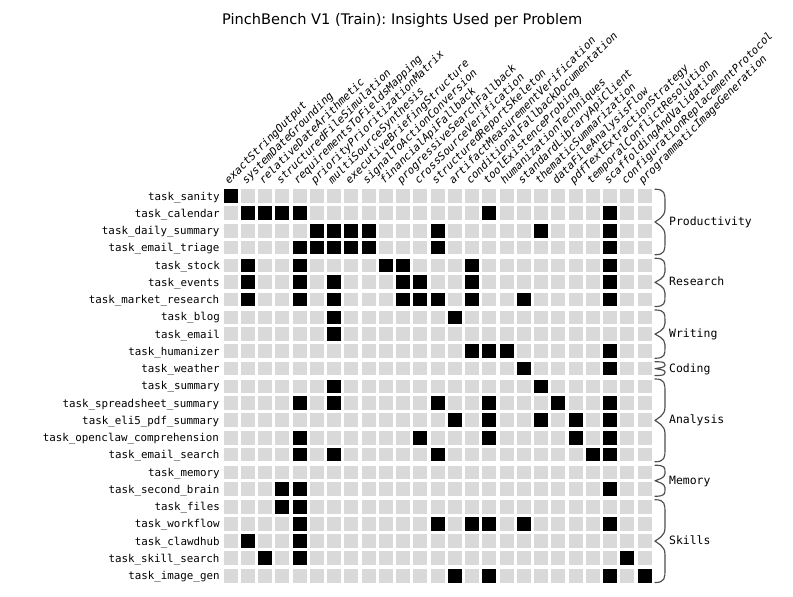}
    \caption{Insight usage across problems in PinchBench V1 (Existing). Different tasks exhibit distinct insight-usage patterns, even within the same task type.}
    \label{fig:wafflepinchbenchv1}
\end{figure}

\begin{figure}[!h]
    \centering
    \includegraphics[width=0.97\linewidth]{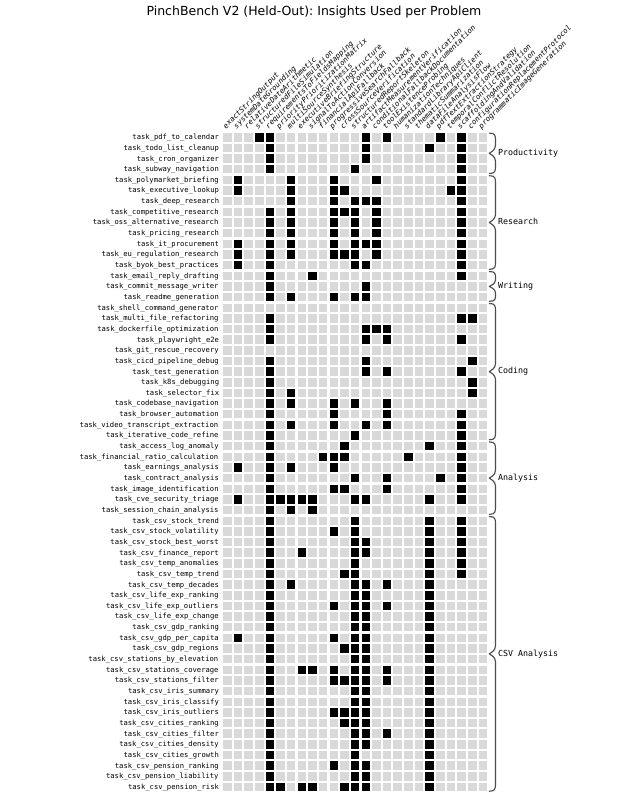}
    \caption{Insight usage across problems in PinchBench V2 (Held-Out). Tasks within the same task type sometimes share similar insight-usage patterns, but a single template cannot capture all patterns within a category, and 2--3 personalized skills are often insufficient.}
    \label{fig:wafflepinchbenchv2}
\end{figure}

\begin{figure}[!h]
    \centering
    \includegraphics[width=0.97\linewidth]{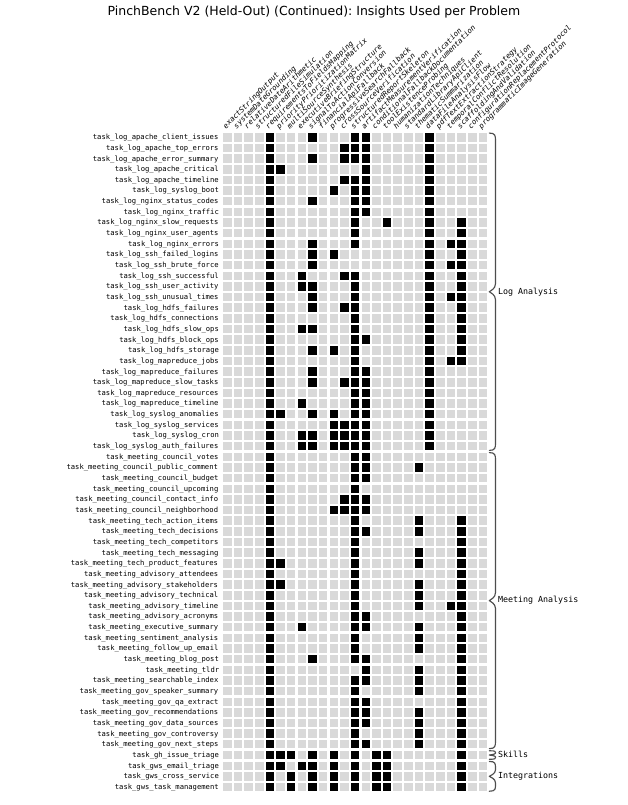}
    \caption{\Cref{fig:wafflepinchbenchv2} (continued)}
    \label{fig:wafflepinchbenchv2continue}
\end{figure}

\begin{figure}[!h]
    \centering
    \includegraphics[width=0.97\linewidth]{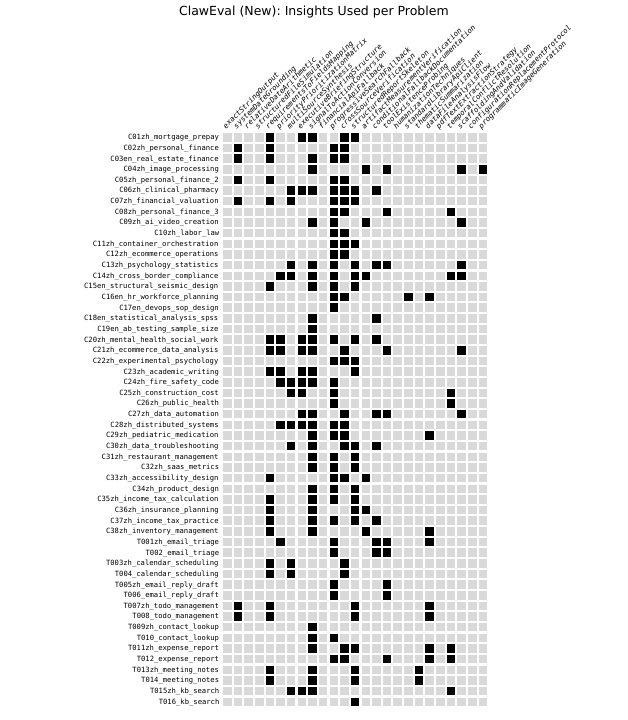}
    \caption{Insight usage across problems in ClawEval (New). We observe similar patterns, while some insights are unused by held-out and new tasks and remain specific to existing tasks.}
    \label{fig:waffleclaweval}
\end{figure}

\begin{figure}[!h]
    \centering
    \includegraphics[width=0.97\linewidth]{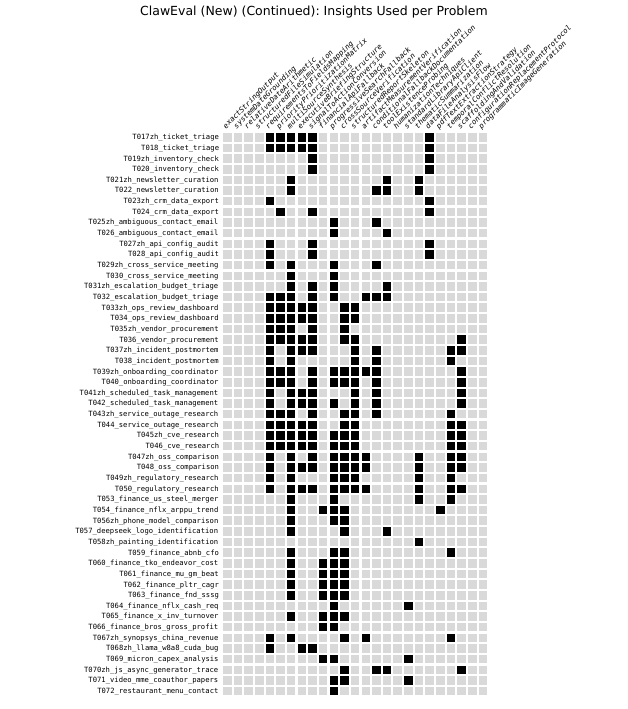}
    \caption{\Cref{fig:waffleclaweval} (continued)}
    \label{fig:waffleclawevalcontinue}
\end{figure}

Different problems, even within the same or closely related task types, select different combinations of insights best suited to their needs. Thus, assigning a fixed template to each cluster or task type is often insufficient, as personalization requirements vary even within a category. For example, the ``integrations'' task type uses only two insight combinations, whereas ``log analysis'' requires more than ten. FoT therefore provides finer-grained and more flexible personalization.

\clearpage
\section{Complete Prompts}
\label[appendix]{sec:prompt}

\subsection{Prompts for Our Instantiation of FoT} 
\Cref{prompt:solution} shows the prompt used by agents when an insight library is available. In \Cref{sec:openclaw}, we incorporate this prompt into the OpenClaw system prompt by specifying that ``available insights are stored in \texttt{insight.md}''. When no insight library is provided (i.e., isolated agents), the prompt reduces to ``Problem: \{problem\}'', where \{problem\} corresponds to the original dataset query. We list each problem prompt in \Cref{sec:problem}.

\captionof{prompt}{Generating solution}
\label{prompt:solution}
\begin{lstlisting}[style=promptbox]
Available Insights to Guide Your Solution:

{insight library}

---
INSTRUCTIONS: Review the insights above and actively apply the relevant techniques from insights to solve this problem. Consider which insights can help you approach the problem more effectively.

Problem: {problem}
\end{lstlisting}

\Cref{prompt:reflection} is used for solution reflection, which is the first part of `Step 2: Local reasoning and self-reflection' (\Cref{fig:workflow}), before an agent generates reasoning traces.

\captionof{prompt}{Reflecting on Solution}
\label{prompt:reflection}
\begin{lstlisting}[style=promptbox]
Analyze the solution below to extract procedural knowledge that reflects the reasoning traces.

Problem:
{problem}

Step-by-Step Solution:
{solution}

Your task: Extract the fundamental techniques used in resolution that can be packaged as reasoning traces. Focus on:

1. What step-by-step procedures were used? How can these be repeated?
2. What conditions determined which approach to use? When should each technique apply?
3. What methods, strategies, or workflows can be applied to similar problems?
4. What made this approach effective? What should someone know to use it correctly?
5. What types of problems would benefit from these techniques?

Output your analysis covering:

### I. Procedural Knowledge
- Break down the solution into clear, repeatable procedures
- The extracted traces should be concrete solutions rather than general principles.

### II. Reusable Techniques and Methods
- List specific techniques, strategies, or workflows used
- The techniques should be solid on practical questions rather than very general and high-level principles.
- For each technique, identify:
  * When it should be used (conditions/triggers)
  * How it was applied (concrete steps)
  * Why it was effective (insights)
  * What problems it could solve (applicability)

### III. Critical Insights and Guidelines
- What key insights made this solution work?
- What common pitfalls should be avoided?
- What variations or edge cases should be considered?

Focus on extracting actionable, procedural knowledge that can be packaged as reusable insights for similar problems.
\end{lstlisting}

\Cref{prompt:insight} is used for the generation of reasoning traces, which is the second part of `Step 2: Local reasoning and self-reflection' (\Cref{fig:workflow}). The description includes our definitions of reasoning traces.

\captionof{prompt}{Generating Reasoning Trace}
\label{prompt:insight}
\begin{lstlisting}[style=promptbox]
Extract reasoning traces from the solution below. Analyze the solution and reflection to identify concrete, actionable traces so that similar problems can be solved via the traces

Problem: {problem}

Solution: {solution}

Reflection: {reflection}

**Your Task:**
Identify and extract all reusable reasoning traces, techniques, and methods used in the solution. Each trace should be a concrete procedure that can guide someone to solve similar problems.

**What Makes a Good Reasoning Trace:**
- A specific technique or method that was used in the solution.
- Something that can be applied to similar problems, not just this one.
- Includes guidance on when and how to use it with clear steps that can be followed if necessary.
- Not repetition of already well-known or commonly adopted techniques.
- Not too general and high-level but contains actionable procedural knowledge.

**Description Must Include:**
1. **Core idea**: The fundamental concept of what this trace is about. What is the main technique or method? What does it do?

2. **When to use**: Explain when this skill should be applied. What types of problems? What conditions must be met? What situations trigger this skill?

**Output Format (Simple JSON):**
Output a simple JSON object with skill names as keys and descriptions as string values:

{{"trace_name": "description"}}

Format Rules:
- Use valid JSON format
- Each trace name must start with "trace_"
- Keep JSON simple - no nested objects, just key-value pairs
- Escape quotes in descriptions with backslash: \\"

**Example:**
{{
  "trace polynomialFactoring": "The major idea is how we can turn a polynomial into a product of simpler expressions. This skill is particularly useful for quadratic and higher-degree polynomial equations where factoring can simplify the problem. Factoring reduces complex polynomials to simpler equations. When solving equations with polynomial expressions that can be factored, especially when the polynomial has recognizable patterns like difference of squares (a^2-b^2), perfect square trinomials (a^2+2ab+b^2), or common factors.  
  "trace depthFirstSearchImplementation": "This algorithm is essential for problems involving path finding, cycle detection, topological sorting, connected components, or exploring all possible solutions in a search space. DFS explores depth before breadth, using stack-based recursion or explicit stack. It is memory-efficient for deep structures and naturally handles backtracking. The visited set prevents infinite loops and redundant work. DFS is the foundation for many graph algorithms including topological sort, strongly connected components, and maze solving. When you need to explore or traverse a graph, tree, or nested structure systematically, going as deep as possible before backtracking. Use DFS when you need to visit all nodes in a connected component, find paths between nodes, detect cycles, or explore recursive structures like file systems, nested data, or game states. "
}}

**Output your response as a valid JSON object only:**
\end{lstlisting}

\Cref{prompt:relation} is used for clustering and identifying connections between reasoning traces, as part of `Step 4: Aggregate traces to update the library' in \Cref{fig:workflow}, before updating the insight library.

\captionof{prompt}{Constructing Relationship between Reasoning Traces}
\label{prompt:relation}
\begin{lstlisting}[style=promptbox]
You are analyzing a collection of reasoning traces generated for problem-solving.  Understand their relationships and structure and build a profile of their relationships.

**Collected {len(insight_store)} traces in total:**
Note: This collection includes ALL traces from all problems without deduplication. Similar names with different indices (e.g., _001, _002) represent different occurrences that may have variations in their descriptions.

{insights_text}

**Your Task:**
Analyze these traces and build a profile of their relationships:

1. **Identify Clusters**:
   Group related traces that share:
   - Resolve the same or similar problem
   - Similar approaches or techniques
   - Nearly identical traces (e.g., same trace with minor variations in description or parameters)
   - Traces in the same cluster should be highly similar.

2. **Build Trace Relationships**:
Record all important relationships - traces don't exist in isolation and build a relationship graph that records:
   - **Prerequisite relationships**: Traces that must be learned/used before others
   - **Composition relationships**: Traces that can be chained/composed together
   - **Alternative relationships**: Different approaches to the same problem
   - **Complementary relationships**: Traces that work better together than individually used
   - **Derivation relationships**: Traces derived from or based on others
   - **Similar relationships**: Traces that are similar but not identical
Map relationships between traces within clusters and across clusters.

# Output Format:
{{
  "clusters": [
    {{
      "cluster_id": 0,
      "cluster_name": "Domain/Theme Name",
      "traces": ["name1", "name2", "name3"],
      "theme": "What is the high-level technical idea of the traces in this cluster?",
    }}
  ],
  "relationships": [
    {{
      "trace_a": "trace_name1",
      "trace_b": "trace_name2",
      "relationship_type": "prerequisite/complementary/alternative/similar/derived_from/
      composes_with",
      "description": "How these traces relate to each other and Why"
    }}
  ]
}}

**Output your analysis as JSON only:**
\end{lstlisting}

The output relationship dictionary from \Cref{prompt:relation} is double-checked and formatted as valid JSON to facilitate downstream extraction and processing. \Cref{prompt:encyclopedia} is used for updating the insight library, as the second part of `Step 4: Aggregate traces to update the library' in \Cref{fig:workflow}. 

\captionof{prompt}{Building Insight Library}
\label{prompt:encyclopedia}
\begin{lstlisting}[style=promptbox]
**Your Task:**
You are extracting fundamental insights from a collection of problem-solving traces.

**Output Requirements (STRICT):**
1. Return EXACTLY one valid JSON object and nothing else.
2. Do NOT output markdown code fences.
3. Do NOT output explanations, notes, reasoning, prefixes, suffixes, or `<think>` content.
4. Do NOT output list/array at top-level.
5. Every key must start with "insight_".
6. Every value must be a single string.
7. No nested objects, no nested arrays.

**Required JSON shape:**
{{
    "insight_name1": "description string",
    "insight_name2": "description string"
}}

**Formatting Rules:**
- Use valid JSON syntax only.
- Keep top-level as key-value pairs only.
- Escape quotes in descriptions with backslash: \\" 
- If uncertain, still output a valid JSON object (possibly with fewer insights), never free text.

Your goal is to extract a comprehensive set of fundamental, cross-domain insights that can be derived and applied beyond their original domain and meet the following requirements: 
- Combine previous insights (if any): {existing_encyclopedia if existing_encyclopedia else "None"} with new insights.
- Extract your insights based on all client reasoning traces: {all_insights_text}. These traces are derived from solving specific problems (bottom-up approach)
- Use clusters of reasoning traces: {clusters_text if clusters_text else "None identified"} to help organize.
- Use relationships between traces: {json.dumps(profiling.get('relationships', []), indent=2) if isinstance(profiling, dict) else "None identified"} to help organization
- Your task is to extract multi-disciplinary, fundamental knowledge (top-down approach) which can be generalized to multi-domain problem-solving.
- The extracted insights should be able to DERIVE and GUIDE the use of the collected insights
- The extracted insights cannot be too general. They are not supposed to be knowledge which can be applied to any problem. They should be fundamental knowledge to particular several domains but specific.
- You should extract proper number of insights. Not too few. Not too many. Do not over simplify or be too detailed.
- DO NOT over-merge insights.

Insights should have following properties:
1. **Extract Reusable Primitives**:
   - For EACH cluster, extract multiple fundamental insights capturing core essence and variations (DO NOT over-merge)
   - Identify cross-domain patterns that apply to multiple fields
   - Create reusable, composable primitives specific enough to be actionable

2. **Knowledge to include**:
   - **Fundamental Level**: Core principles underlying multiple domains
   - **General Level**: Broad techniques for related problem types
   - **Cross-Domain**: Insights transferable beyond origin field

3. **Preserve While Generalizing**:
   - Create fundamental versions that can guide/derive original insights
   - Maintain important variations rather than collapsing into single insight

4. **Description Format:**
   Each description is a single string containing:
   - What the insight is and how it solves problems
   - When to use: problem types, conditions, triggers (be comprehensive and specific)

**Example 1 - Transformer Architecture:**

Input reasoning traces:
- reasoning trace VisionTransformerImageClassification: "I need to classify medical X-ray images into disease categories. CNNs aren't working well - they can't capture relationships between distant regions like how fluid in the lower right lung might relate to heart enlargement. Let me try Vision Transformer (ViT). I'll divide each X-ray into 16x16 patches - a 224x224 image gives 196 patches. Each patch becomes like a token in NLP. I flatten each to a 256-dim vector. Since transformers don't know spatial positions, I add position embeddings so the model knows patch [0,0] is top-left. I prepend a [CLS] token to gather global info. Feeding through 12 transformer encoder layers - the self-attention lets every patch attend to every other patch directly, so patch [2,5] can look at patch [10,12] even though they're far apart spatially. This is exactly what I need! After 12 layers, I extract the [CLS] token and feed to MLP classifier. Training on 50k chest X-rays: 94.2% accuracy, beating ResNet-50's 89.3%. The attention maps show it's correctly attending to both lungs simultaneously for pneumonia detection, linking heart size to lung fluid - this cross-region reasoning is what CNNs miss. The key: treating image patches as tokens with self-attention enables global spatial reasoning."

- reasoning trace TransformerNextWordPrediction: "I'm building autocomplete for a text editor. The challenge: predict next word given arbitrary-length context. RNNs struggle with long sequences - the hidden state forgets earlier context. Let me use a transformer decoder. I tokenize 'The cat sat on the' using WordPiece -> tokens [254, 8901, 4523, 651, 278]. Convert each to 512-dim embedding and add positional encodings. Critical part: causal masking so the model can't cheat by seeing future words. When predicting token at position 4, it should only see 0-3. I implement lower-triangular attention mask. Processing through 6 decoder layers with masked self-attention. At position 4 ('the'), attention computes similarity between its query and keys of previous words. It attends strongly to 'sat' (0.8) and 'on' (0.7), weakly to 'cat' (0.2). Using 12 heads helps - different heads capture different patterns: head-2 learns syntax (preposition+article), head-5 learns semantics (actions+objects), head-8 learns long-range dependencies. After final layer, project last hidden state through 50k-dim softmax. For 'the': top predictions are 'mat' (0.73), 'floor' (0.12), 'rug' (0.08). Deployed in production - users accept top-3 suggestion 85\% of the time, reducing typing by 40\%. Transformer's self-attention captures context way better than RNN's sequential processing."

Output aggregated insight:
{{
  "insight_transformerArchitecture": "This fundamental neural network architecture applies across natural language processing, computer vision, time series analysis, graph neural networks, and multi-modal learning. This insight is essential for modern AI applications including language models, image processing, code generation, and scientific computing. When you need to capture relationships between all elements simultaneously (self-attention), you're working with sequences of variable length, you need parallel processing of sequences, or when the problem involves understanding context and relationships. Details: 1) Design input representation - convert your data into embeddings (token embeddings for text, patch embeddings for images, node embeddings for graphs), add positional encodings to preserve sequence information, and prepare input for transformer blocks 2) Create models with transformer blocks 3) Apply task-specific architecture - use encoder-only for understanding (BERT, ViT), decoder-only for generation (GPT), or encoder-decoder for translation"
}}

**Example 2 - Surface-Enhanced Raman Spectroscopy:**

Input reasoning traces:
- reasoning trace SERSMedicalDetectionR6G: "I need to detect cancer biomarkers in blood at incredibly low concentrations - 10^-12 M, like finding molecules in a swimming pool. ELISA only goes to 10^-9 M, not sensitive enough for early diagnosis. Let me try SERS - Surface-Enhanced Raman Spectroscopy. Metal nanoparticles create huge EM field enhancements. I synthesize 60nm gold nanoparticles via citrate reduction. At 785nm laser, these have plasmon resonance amplifying local field by ~10^6. But I need selectivity too - can't detect everything. So I functionalize the gold with anti-PSA antibodies for prostate cancer. When I add patient serum, only PSA proteins bind. Here's the clever part: I add R6G (Rhodamine 6G) reporter molecules. R6G has enormous Raman cross-section and when it sits in nanogaps between gold particles, field enhancement shoots to 10^8 or 10^10. Incubate 30 min for PSA binding, add R6G which sticks near bound PSA. Hit with 785nm laser at 5mW - I see characteristic R6G peaks at 1650, 1510, 1310 cm^-1. Peak intensity directly proportional to PSA amount. Integrate 60 sec for good SNR. Comparing to calibration: detecting PSA at 0.1 ng/mL - that's 10,000x more sensitive than ELISA! On clinical samples, detected prostate cancer 3-6 months earlier than conventional tests. The breakthrough: combining selective antibody recognition with SERS amplification gives single-molecule sensitivity while maintaining specificity."

- reasoning trace SERSPollutantDetection: "Monitoring river water for pesticides. EPA limit for malathion is 0.1 ppb but standard chromatography needs 1 ppb minimum. I need 10x better for early warning. SERS might work. Instead of spherical particles, I'll fabricate silver nanorod arrays - sharp tips create hotter hotspots than spheres. Using oblique angle deposition: 80nm nanorods with 4:1 aspect ratio on silicon. Gaps between rods only 5-10nm - perfect for trapping molecules. I calculate enhancement should hit 10^10 at 532nm. Collect river water, filter through 0.2\mu m to remove debris and bacteria. Drop 50 \mu L onto nanorod substrate. During 5-min adsorption, pesticide molecules diffuse into nanogaps. Small gap means molecules guaranteed in enhancement zone (<10nm from metal). Rinse gently - removes interfering organics/salts but leaves adsorbed pesticides. Excite with 532nm at 2mW, matching silver plasmon peak. Even at 0.01 ppb malathion, clear peaks at 1440 cm^-1 (P=S stretch), 1080 cm^-1 (P-O-C), 640 cm^-1 (C-S). Measuring 1440 peak height vs calibration standards for quantification. Tested 50 river sites, cross-validated against LC-MS: R^2=0.97 correlation. Best part: do this in field with portable Raman - no lab needed. Real-time monitoring at 10x below regulatory limits. The nanorod geometry is critical - those sharp tips and tight gaps push enhancement to 10^10."

Output aggregated insight:
{{
  "insight_surfaceEnhancedRamanSpectroscopy": "This powerful technique applies across analytical chemistry, materials science, biosensing, pharmaceutical analysis, environmental monitoring, and forensics. The technique achieves single-molecule sensitivity (10^6-10^11 enhancement) while providing molecular structural information through vibrational fingerprints. When to use: When you need ultra-sensitive detection below conventional analytical limits, when you want label-free molecular identification, when analyzing trace contaminants or biomarkers, or when field-portable real-time analysis is required. Common steps: 1) Prepare SERS-active substrate - synthesize plasmonic nanostructures (gold/silver nanoparticles, nanorods, nanostars) optimizing particle size (20-100 nm), shape, and inter-particle spacing (1-10 nm gaps) to maximize electromagnetic field enhancement at laser wavelength 2) Functionalize substrate if needed - modify metallic surface with antibodies, aptamers, or molecular recognition elements for selective analyte binding and improved specificity 3) Prepare and apply sample - process sample (filter, dilute, concentrate as needed), deposit onto SERS substrate via drop-casting or flow-through, allow adsorption time for molecules to enter hot spots (<10 nm from metal surface) 4) Select laser parameters - choose wavelength matching plasmon resonance (532, 633, or 785 nm), optimize power (0.1-10 mW) to avoid sample damage while maximizing signal 5) Acquire SERS spectrum - collect Raman scattered light with appropriate integration time, record vibrational spectrum showing characteristic molecular peaks 6) Analyze spectral fingerprint - identify molecules by comparing peak positions to reference spectra, quantify concentration from peak intensities using calibration curves, assess molecular orientation from peak ratios 7) Validate and control quality - average multiple spots for reproducibility, use internal standards, verify with orthogonal methods, consider substrate heterogeneity and enhancement factor variations"
}}
\end{lstlisting}

\Cref{prompt:criteria} is used for determining whether a paper can be covered by insights from the library, in the research insight discovery application.

\captionof{prompt}{Judging Whether Paper Contributions Match with the Insight Library}
\label{prompt:criteria}
\begin{lstlisting}[style=promptbox]
You are evaluating whether a research paper's proposed solutions are guided by or directly derived from insights in an encyclopedia. You should focus primarily on the content of the paper and the insights, while relying as little as possible on external knowledge. For example, you may have read the paper before or already know the core techniques discussed in the insight or the paper. However, conduct the evaluation as if you did not know them beforehand, except for very basic machine learning knowledge.

Insights:
{insights_prompt}

Research Paper:
{paper_text}

Evaluation Criteria - An insight guides the paper ONLY IF ALL of the following are true:

1. CONCRETE METHODOLOGY USAGE: The insight's methodology or approach is concretely used in the paper's methods/approach section, not just theoretically relevant or mentioned in motivation.

2. METHODS SECTION PRESENCE: The insight must be related to how the paper actually implements its solution (methods, algorithms, techniques), not just in problem statement or related work.

3. SPECIFICITY: The insight must specifically address a key challenge or component of the paper's solution, not just be generally applicable background knowledge.

Response Format:
- Respond ONLY in valid JSON with keys: guided (boolean), matched_insights (array of insight names)
- Set guided=true ONLY when at least one insight passes ALL criteria above
- Use only exact insight names from the Insights list above
\end{lstlisting}

\subsection{Complete Problem Prompts}
\label[appendix]{sec:problem}
For completeness, we summarize the problem prompts used in each application. For the research insight library, we adopt the template in \Cref{prompt:paperreading}, using the answer as the ``\{solution\}.'' For PinchBench and ClawEval, we directly use the task instructions provided by the benchmarks.

\captionof{prompt}{Humanity's Last Exam}
\label{prompt:hle}
\begin{lstlisting}[style=promptbox]
You are solving one HLE multimodal benchmark problem.
Use BOTH the question text and the attached image.
Return only the final answer, with no explanation.
\end{lstlisting}

\captionof{prompt}{AIME26}
\label{prompt:math}
\begin{lstlisting}[style=promptbox]
'\nSolve the problem step by step. Wrap your final answer in "\\boxed{}".'
\end{lstlisting}

\captionof{prompt}{GPQA Diamond}
\label{prompt:science}
\begin{lstlisting}[style=promptbox]
Please solve this graduate-level science question step by step.

Instructions:
1. Analyze the question carefully
2. Consider each option systematically
3. Explain your reasoning
4. Provide your final answer as a single letter (A, B, C, or D)
5. Wrap your final answer letter in \\boxed{}, for example: \\boxed{A}
\end{lstlisting}

\captionof{prompt}{LiveCodeBench V6}
\label{prompt:coding}
\begin{lstlisting}[style=promptbox]
Please solve this programming problem by writing clean, efficient code.

Requirements:
1. Read the problem description carefully
2. Understand the input/output format
3. Write a complete solution
4. Include proper input/output handling
5. Wrap your final code solution in markdown code blocks with triple backticks (```)

Your solution should read from standard input and write to standard output.
\end{lstlisting}

\captionof{prompt}{Paper Reading}
\label{prompt:paperreading}
\begin{lstlisting}[style=promptbox]
Answer research question of the paper: {paper_name} based on paper content:{paper_content}
All papers need to be considered in the analysis. 
\end{lstlisting}

\subsection{Prompts for Metacognitive Reuse}
\label[appendix]{apx:bciprompt}
For completeness, here, we re-state the prompts proposed in prior work~\citep{didolkar2025metacognitive} that we build upon and compare against.
\captionof{prompt}{Generating Solution}
\begin{lstlisting}[style=promptbox]
Please reason step by step and put the final answer in
Problem: <problem> 
\end{lstlisting}

\captionof{prompt}{Generating Reflection}
\begin{lstlisting}[style=promptbox]
Here is the definition of a behavior:
- A behavior is a note or skill to keep in mind while solving math problems.
- It can be a strategy, a trick, or a technique.
- It can also be a general rule or a common sense principle.
- A behavior is not a solution to the problem, but it can be used to solve the problem.

For example - if the problem is "Find the area of a circle with radius 4", one useful behaviour could be
{"behavior_area_of_circle": area of a circle is pi*r^2}.

Given a problem and the corresponding solution, react and critique the solutions along the following dimensions:

1. Correctness Analysis: Is the answer mathematically correct? Are there calculation errors? Is the reasoning logically
sound? Are all steps properly justified? What specific mistakes were made?
2. Missing Behaviors Analysis: What behaviors should have been used but weren't? Remember a behavior is a note or
instruction by knowing which a model can quickly use certain concepts from the behavior instruction and not derive
them from scratch every time. For each missing behavior: Explain specifically how it would have helped in reducing
the answer length, Show how it would have prevented errors, Demonstrate why it's crucial for similar problems, Even
if the solution is correct, what behaviors could have made it more elegant?
3. New Behavior Suggestions: Suggest specific new behaviors that will help with similar problems. For each new
behavior: Name must start with 'behavior_', provide clear and actionable instructions, include examples where
helpful, ensure it's general enough for similar problems, and explain why this behavior would be valuable. 
\end{lstlisting}

\captionof{prompt}{Generating Behavior}
\begin{lstlisting}[style=promptbox]
<reflection prompt>

<reflection>

Now, given this reflection generate a list of
behaviors and corresponding instructions explanations in json format. Each behavior should be a single line, and the format is "behavior_[name]: [description]". The list should
be in json format, and each behavior should be a key-value pair, where the key is the behavior name and the value is the description. 
\end{lstlisting}

\section{Raw Reasoning Trajectories and Reasoning Traces}

\subsection{Complete Raw Reasoning Process for Case Studies in \Cref{sec:supplymultidomain}}
\label[appendix]{appx:completereasoning}

{\renewcommand{\lstlistingname}{Reasoning Process
}\setcounter{lstlisting}{0}

\captionof{reasoningprocess}{Incorrect Reasoning Process of AIME26 Problem 20}
\label{reasonaime26p20incorrect}
\begin{lstlisting}[style=promptbox]
**Exploring Marble Ratios**

I'm currently focused on understanding the interplay between the number of red ($R$) and blue ($B$) marbles, given the constraints $R \ge 7$ and $B \ge 7$. My thoughts are beginning to consider scenarios involving drawing a specific quantity ($k$) of marbles from this urn.

**Deriving Marble Counts**

My current focus is on simplifying the equality derived from the hypergeometric probabilities. I've expanded the binomial coefficients and am now working through the algebraic cancellations to isolate the relationship between $R$ and $B$.

GCD(144, 240) = $2^4 \times 3 = 16 \times 3 = 48$.

Divide both sides by 48:
$5(B-2) = 3(R-4)$.
$5B - 10 = 3R - 12$.
$5B = 3R - 2$.

This equation relates the number of red marbles ($R$) and blue marbles ($B$).
We are given $R \ge 7$ and $B \ge 7$.
We need to find integer solutions $(R, B)$ that satisfy $5B = 3R - 2$ and these minimum conditions.

Let's express $R$ in terms of $B$:
$3R = 5B + 2$.
$R = \frac{5B + 2}{3}$.

Since $R$ must be an integer, $5B + 2$ must be divisible by 3.
We can test values of $B \ge 7$:
If $B=7$, $R = \frac{5(7)+2}{3} = \frac{37}{3}$, not an integer.
If $B=8$, $R = \frac{5(8)+2}{3} = \frac{42}{3} = 14$. This is an integer.
Let's check the conditions: $R=14 \ge 7$ and $B=8 \ge 7$. This is a valid solution.

Let's test further values of $B$ to see if there are other solutions.
For $5B + 2$ to be divisible by 3, we can look at the remainders of $B$ when divided by 3.
If $B \equiv 0 \pmod{3}$: $5(0) + 2 = 2 \pmod{3}$. Not divisible.
If $B \equiv 1 \pmod{3}$: $5(1) + 2 = 7 \equiv 1 \pmod{3}$. Not divisible.
If $B \equiv 2 \pmod{3}$: $5(2) + 2 = 12 \equiv 0 \pmod{3}$. Divisible.

So, $B$ must be of the form $3m + 2$ for some integer $m$.
We are given $B \ge 7$.
If $m=0$, $B=2$, too small.
If $m=1$, $B=5$, too small.
If $m=2$, $B=8$. This gives $R = \frac{5(8)+2}{3} = 14$. $(14, 8)$ is a solution.
If $m=3$, $B=11$. $R = \frac{5(11)+2}{3} = \frac{57}{3} = 19$. $(19, 11)$ is a solution.
If $m=4$, $B=14$. $R = \frac{5(14)+2}{3} = \frac{72}{3} = 24$. $(24, 14)$ is a solution.

In general, if $B = 3m+2$ and $B \ge 7$, then $m \ge 2$.
$R = \frac{5(3m+2)+2}{3} = \frac{15m+10+2}{3} = \frac{15m+12}{3} = 5m+4$.

So, the solutions for $(R, B)$ are of the form $(5m+4, 3m+2)$ where $m \ge 2$ is an integer.
The smallest solution is when $m=2$: $(5(2)+4, 3(2)+2) = (14, 8)$.
This confirms our earlier findings.

The question asks for the ratio of red to blue marbles.
The equation is $5B = 3R - 2$.

If the question is asking for *a* ratio, we can use any valid pair.
For $(14, 8)$, the ratio $R:B = 14:8 = 7:4$.
For $(19, 11)$, the ratio $R:B = 19:11$.
For $(24, 14)$, the ratio $R:B = 24:14 = 12:7$.

Let's re-examine the derivation to ensure no steps were missed.
The initial equation was $\binom{R}{4} \binom{B}{3} = \binom{R}{5} \binom{B}{2}$.
This simplified to $\frac{R-4}{5} \frac{B-2}{3} = 1$. (This is a common shortcut for relating binomial coefficients).
Let's check this shortcut:
$\binom{R}{k} \binom{B}{l} = \binom{R}{k+1} \binom{B}{l-1}$
$\frac{\binom{R}{k}}{\binom{R}{k+1}} = \frac{\binom{B}{l-1}}{\binom{B}{l}}$
$\frac{R!/(k!(R-k)!)}{R!/((k+1)!(R-k-1)!)} = \frac{B!/((l-1)!(B-l+1)!)}{B!/(l!(B-l)!)}$
$\frac{(k+1)!(R-k-1)!}{k!(R-k)!} = \frac{l!(B-l)!}{(l-1)!(B-l+1)!}$
$\frac{k+1}{R-k} = \frac{l}{B-l+1}$

In our case, $k=4$ and $l=3$.
$\frac{4+1}{R-4} = \frac{3}{B-3+1}$
$\frac{5}{R-4} = \frac{3}{B-2}$
$5(B-2) = 3(R-4)$. This matches our derivation.

The equation $5B = 3R - 2$ is the core relationship.
The problem implies a single answer for the ratio. This usually means there is a unique solution or a specific structure that defines the ratio.

The relationship $5B = 3R - 2$ can be rewritten as $3R - 5B = 2$.
This is a linear Diophantine equation.
We found a particular solution $(R_0, B_0) = (14, 8)$.
The general solution is given by:
$R = R_0 + \frac{-5}{gcd(3, -5)} t = 14 + \frac{-5}{1} t = 14 - 5t$
$B = B_0 - \frac{3}{gcd(3, -5)} t = 8 - \frac{3}{1} t = 8 - 3t$
where $t$ is an integer.

However, we have the constraints $R \ge 7$ and $B \ge 7$.
$14 - 5t \ge 7 \implies 7 \ge 5t \implies t \le \frac{7}{5}$. So $t \le 1$.
$8 - 3t \ge 7 \implies 1 \ge 3t \implies t \le \frac{1}{3}$. So $t \le 0$.

Combining these, $t \le 0$.
If $t=0$: $R = 14$, $B = 8$. This gives $R:B = 14:8 = 7:4$.
If $t=-1$: $R = 14 - 5(-1) = 19$, $B = 8 - 3(-1) = 11$. This gives $R:B = 19:11$.
If $t=-2$: $R = 14 - 5(-2) = 24$, $B = 8 - 3(-2) = 14$. This gives $R:B = 24:14 = 12:7$.

Let's recheck the common simplification for binomial coefficients.
$\binom{n}{k} = \frac{n-k+1}{k} \binom{n}{k-1}$.
So, $\binom{R}{5} = \frac{R-5+1}{5} \binom{R}{4} = \frac{R-4}{5} \binom{R}{4}$.
And $\binom{B}{3} = \frac{B-3+1}{3} \binom{B}{2} = \frac{B-2}{3} \binom{B}{2}$.

Our original equation: $\binom{R}{4} \binom{B}{3} = \binom{R}{5} \binom{B}{2}$.
Substitute the relations:
$\binom{R}{4} \left( \frac{B-2}{3} \binom{B}{2} \right) = \left( \frac{R-4}{5} \binom{R}{4} \right) \binom{B}{2}$.

Assuming $\binom{R}{4} \neq 0$ and $\binom{B}{2} \neq 0$ (which is true for $R \ge 7, B \ge 7$ and the context of drawing marbles), we can cancel them out.
$\frac{B-2}{3} = \frac{R-4}{5}$.
$5(B-2) = 3(R-4)$.
$5B - 10 = 3R - 12$.
$5B = 3R - 2$.

This equation is correct.
The problem is that there are multiple integer solutions for $(R,B)$ that satisfy $R \ge 7$ and $B \ge 7$.
The phrasing "the ratio of red to blue marbles" implies a unique ratio.

Is there any other constraint I missed?
"The probability that exactly 4 of them are red equals the probability that exactly 5 of them are red."
This is what we used.

What if the question implies the *simplest* integer ratio derived from the fundamental relationship?
The equation $3R - 5B = 2$ is the fundamental relationship.

If we consider the equation modulo 3:
$-5B \equiv 2 \pmod{3}$
$B \equiv 2 \pmod{3}$.
This means $B$ can be $2, 5, 8, 11, 14, \ldots$.

If we consider the equation modulo 5:
$3R \equiv 2 \pmod{5}$.
Multiply by 2 (inverse of 3 mod 5):
$6R \equiv 4 \pmod{5}$
$R \equiv 4 \pmod{5}$.
This means $R$ can be $4, 9, 14, 19, 24, \ldots$.

Combining the conditions $R \equiv 4 \pmod{5}$ and $B \equiv 2 \pmod{3}$:
We need solutions for $(R,B)$ that satisfy $3R - 5B = 2$.
We found $(14, 8)$ as a solution.
$14 \equiv 4 \pmod{5}$ (correct)
$8 \equiv 2 \pmod{3}$ (correct)

Let's look at the structure of the ratio $R/B$ from our general solutions $(5m+4, 3m+2)$ for $m \ge 2$.
$m=2: (14, 8) \implies R/B = 14/8 = 7/4$.
$m=3: (19, 11) \implies R/B = 19/11$.
$m=4: (24, 14) \implies R/B = 24/14 = 12/7$.

The ratios are not constant. This means my interpretation that there's a single ratio might be wrong, or there's an implicit assumption in the problem statement that leads to a unique ratio.

Could the problem be seeking the ratio derived directly from the equation $5B = 3R - 2$ *before* considering the constraints $R \ge 7, B \ge 7$?
If we rearrange $5B = 3R - 2$ to form a ratio, it's not straightforward.

Let's go back to the simplified equation:
$\frac{B-2}{3} = \frac{R-4}{5}$.
This shows a relationship between the *deviations* from certain numbers.
Let $R-4 = 5x$ and $B-2 = 3x$.
Then $R = 5x + 4$ and $B = 3x + 2$.
This is the same general solution form we derived for Diophantine equations.
Here, $x$ plays the role of $m$.
So, $R = 5m+4$ and $B = 3m+2$.

We need $R \ge 7$ and $B \ge 7$.
$5m+4 \ge 7 \implies 5m \ge 3 \implies m \ge 3/5$.
$3m+2 \ge 7 \implies 3m \ge 5 \implies m \ge 5/3$.
So, we need $m \ge 5/3$. Since $m$ must be an integer for $R$ and $B$ to be integers (derived from the binomial coefficient formulation, where $m$ represents a parameter in the general solution of the Diophantine equation), the smallest integer value for $m$ is 2.

This leads to the same set of solutions:
$m=2: (R,B) = (14, 8)$. Ratio $14:8 = 7:4$.
$m=3: (R,B) = (19, 11)$. Ratio $19:11$.
$m=4: (R,B) = (24, 14)$. Ratio $24:14 = 12:7$.

Is it possible that the problem is asking for the ratio $R:B$ derived from the equation $3R - 5B = 2$ itself, as if it were $3R = 5B$? This is not mathematically sound.

Consider the structure of the question again. The goal is to find the ratio of red to blue marbles.
The fundamental equation is $5B = 3R - 2$.

Let's re-evaluate the use of "ratio." Typically, a ratio $a:b$ implies a simplified form.
If the question implies that the *proportion* of red to blue marbles is constant, then that would mean $R/B$ is constant.
But we've shown that $R/B$ is not constant for different integer solutions.

Perhaps the problem is asking for the ratio based on the smallest possible integer values of $R$ and $B$ that satisfy the derived equation *and* the initial constraints.
The smallest valid integer pair is $(R,B) = (14, 8)$.
This pair gives the ratio $14:8$, which simplifies to $7:4$.

If the problem had been $3R = 5B$, then $R/B = 5/3$.
But we have $3R - 5B = 2$.

Let's consider the ratio from the equation $\frac{B-2}{3} = \frac{R-4}{5}$.
This implies that $\frac{R-4}{B-2} = \frac{5}{3}$.
This is a ratio of deviations.

What if the question implies a continuous proportion and we are looking for the underlying linear relationship?
If $R = \frac{5B+2}{3}$, then the ratio is $R:B = \frac{5B+2}{3} : B$.
This is not a simple constant ratio.

Could it be that the question is poorly phrased, and it intends to ask for the relationship between $R$ and $B$?
Or, could it be that there's a subtlety in "the ratio of red to blue marbles" that means the fundamental ratio between the coefficients in the simplified equation?

The equation is $5B = 3R - 2$.
If we ignore the '-2' term for a moment, we'd have $5B \approx 3R$, leading to $R/B \approx 5/3$.

Let's consider the context of probability.
Hypergeometric distribution implies discrete counts.

Let's assume the problem expects the simplest integer ratio arising from the first valid solution pair.
The first valid solution pair $(R, B)$ satisfying $R \ge 7, B \ge 7$ and $5B = 3R-2$ is $(14, 8)$.
The ratio $R:B = 14:8 = 7:4$.

This seems to be the most plausible interpretation if a single ratio is expected.
The thoughts are progressing towards solving the Diophantine equation and finding the smallest valid integer solution.

**Final Check:**
Equation: $\binom{R}{4} \binom{B}{3} = \binom{R}{5} \binom{B}{2}$
Simplified to: $5(B-2) = 3(R-4)$
Which is: $3R - 5B = 2$
Constraints: $R \ge 7, B \ge 7$.
We are looking for integer solutions $(R, B)$.
The general solution is $R = 14 - 5t, B = 8 - 3t$.
Applying constraints: $t \le 0$.
For $t=0$, $(R,B) = (14, 8)$. Ratio $14:8 = 7:4$.
For $t=-1$, $(R,B) = (19, 11)$. Ratio $19:11$.
For $t=-2$, $(R,B) = (24, 14)$. Ratio $24:14 = 12:7$.

If the problem implies a single ratio, then it must be the one derived from the minimal valid integer solution.

The process is to derive the equation $3R - 5B = 2$, then find the smallest integer solution pair $(R, B)$ that satisfies $R \ge 7$ and $B \ge 7$, and then state the ratio $R:B$ for that pair.
The smallest pair is $(14, 8)$.
The ratio is $14:8$, which simplifies to $7:4$.

Let's re-read the original prompt to make sure there are no implicit meanings I'm missing.
"Given that the probability that exactly 4 of them are red equals the probability that exactly 5 of them are red." This is precisely what was used.

The problem asks for "the ratio of red to blue marbles." This suggests a single, definitive ratio.
The derivation $5(B-2) = 3(R-4)$ is solid.
The solutions are of the form $(14+5m, 8+3m)$ if we had written $R = 14+5m, B = 8+3m$.
Let's recheck the Diophantine solution general form.
$3R - 5B = 2$.
Particular solution $(14, 8)$.
$3(14) - 5(8) = 42 - 40 = 2$. Correct.
General solution:
$R = 14 + (-5/gcd(3,-5))n = 14 - 5n$
$B = 8 - (3/gcd(3,-5))n = 8 - 3n$.
Where $n$ is an integer.

Constraints: $R \ge 7, B \ge 7$.
$14 - 5n \ge 7 \implies 7 \ge 5n \implies n \le 7/5$, so $n \le 1$.
$8 - 3n \ge 7 \implies 1 \ge 3n \implies n \le 1/3$, so $n \le 0$.
Thus, $n \le 0$.

If $n=0$: $R=14, B=8$. Ratio $14:8 = 7:4$.
If $n=-1$: $R=19, B=11$. Ratio $19:11$.
If $n=-2$: $R=24, B=14$. Ratio $24:14 = 12:7$.

The wording "the ratio" is still problematic if there are multiple valid pairs.
Could it be that the problem expects the ratio in its most fundamental form before constraints?
The equation $5B = 3R - 2$ is the fundamental relationship.

If the context was asking for the ratio of coefficients, it would be $3:5$ or $5:3$.
But the question is about the marbles.

Let's consider the initial equation again:
$\binom{R}{4} \binom{B}{3} = \binom{R}{5} \binom{B}{2}$.
If we rearrange this as:
$\frac{\binom{R}{4}}{\binom{R}{5}} = \frac{\binom{B}{2}}{\binom{B}{3}}$.
$\frac{R!/(4!(R-4)!)}{R!/(5!(R-5)!)} = \frac{B!/(2!(B-2)!)}{B!/(3!(B-3)!)} $
$\frac{5!(R-5)!}{4!(R-4)!} = \frac{3!(B-3)!}{2!(B-2)!}$
$\frac{5}{R-4} = \frac{3}{B-2}$.
This is exactly $5(B-2) = 3(R-4)$.

The relationship is between $R-4$ and $B-2$.
This means $(R-4) : (B-2) = 5:3$.
If the question was asking for the ratio of $(R-4)$ to $(B-2)$, it would be $5:3$.
But it asks for the ratio of $R$ to $B$.

It's common in such problems that the simplest integer solution providing the ratio is expected.
The simplest integer solution satisfying $R \ge 7$ and $B \ge 7$ is $(R,B) = (14, 8)$.
The ratio $R:B$ for this pair is $14:8$, which simplifies to $7:4$.

I am going to assume the question implies the ratio derived from the smallest valid integer pair.

The core of my thought process now is to ensure the derivation of the equation $3R-5B=2$ is accurate and that the interpretation of "the ratio" refers to the one obtained from the smallest valid integer solution.
I've confirmed the equation derivation multiple times.
The ambiguity lies in "the ratio." However, standard practice often points to the simplest derived ratio from the minimal valid solution.

**My current focus is on confirming that the smallest valid integer solution yields the intended ratio.**

I have established the core equation $3R - 5B = 2$ and the constraints $R \ge 7, B \ge 7$.
My analysis shows that the general integer solution is $(R,B) = (14-5n, 8-3n)$ for integer $n \le 0$.
The smallest valid integer pair occurs at $n=0$, yielding $(R,B) = (14, 8)$.
The ratio $R:B$ for this pair is $14:8$, which simplifies to $7:4$.
This seems to be the most sensible interpretation of "the ratio" given the problem statement.
Therefore, I am confident that the derived ratio of $7:4$ is correct based on the minimal valid solution.

Update Summary:
My thoughts are solidifying on the derived equation $3R - 5B = 2$.
I've confirmed the smallest integer solution satisfying the constraints $R \ge 7$ and $B \ge 7$ is $(14, 8)$.
The ratio of red to blue marbles derived from this minimal solution is $14:8$, simplifying to $7:4$.

**Refining Marble Ratio**

My focus is on the equation $3R - 5B = 2$ derived from the probability equality. I've confirmed that the smallest integer solution pair satisfying $R \ge 7$ and $B \ge 7$ is $(14, 8)$. This yields the ratio $14:8$, which simplifies to $7:4$.**Solving for Marble Ratio**

My thoughts have converged on the equation $3R - 5B = 2$, derived directly from the problem's probability condition. I have identified the smallest integer solution pair that respects the $R \ge 7$ and $B \ge 7$ constraints as $(14, 8)$. The ratio of red to blue marbles from this minimal pair is $14:8$, simplifying to $7:4$.

**Deriving Marble Equation**

I've successfully derived the core equation $3R - 5B = 2$ from the given probability equality. My focus has now shifted to finding the smallest integer solution pair $(R, B)$ that satisfies the minimum marble count constraints of $R \ge 7$ and $B \ge 7$.**Confirming Minimal Solution**

My current thoughts are focused on the derived equation $3R - 5B = 2$. I have rigorously confirmed that the smallest integer solution pair satisfying the constraints $R \ge 7$ and $B \ge 7$ is $(14, 8)$. This pair leads directly to the ratio $14:8$, which simplifies to $7:4$.**Finalizing Marble Ratio**

I am now confident in the derived equation $3R - 5B = 2$ and have confirmed that the smallest valid integer solution pair respecting $R \ge 7$ and $B \ge 7$ is $(14, 8)$. This yields the ratio of red to blue marbles as $14:8$, simplifying to $7:4$.

**Finalizing Marble Ratio**

I am now confident in the derived equation $3R - 5B = 2$ and have confirmed that the smallest valid integer solution pair respecting $R \ge 7$ and $B \ge 7$ is $(14, 8)$. This yields the ratio of red to blue marbles as $14:8$, simplifying to $7:4$.

**Deriving Marble Ratio**

I have successfully derived the fundamental relationship $3R - 5B = 2$. My current focus is on identifying the smallest integer solution pair $(R, B)$ that satisfies the given constraints $R \ge 7$ and $B \ge 7$. This minimal solution is crucial for determining the definitive ratio.

**Finalizing Marble Ratio**

I am now confident in the derived equation $3R - 5B = 2$ and have confirmed that the smallest valid integer solution pair respecting $R \ge 7$ and $B \ge 7$ is $(14, 8)$. This yields the ratio of red to blue marbles as $14:8$, simplifying to $7:4$.

**Refining Marble Ratio**

My thoughts are solidifying on the derived equation $3R - 5B = 2$. I've confirmed that the smallest integer solution satisfying $R \ge 7$ and $B \ge 7$ is $(14, 8)$, leading to the ratio $14:8$, which simplifies to $7:4$.

**Finalizing Marble Ratio**

My thoughts are now solidified. I've confirmed the core equation $3R - 5B = 2$ and rigorously verified that the smallest integer solution pair satisfying the constraints $R \ge 7$ and $B \ge 7$ is $(14, 8)$. This pair directly yields the ratio of red to blue marbles as $14:8$, which simplifies to $7:4$.

**Confirming Minimal Solution**

My thoughts have converged on the equation $3R - 5B = 2$, derived directly from the problem's probability condition. I have identified the smallest integer solution pair that respects the $R \ge 7$ and $B \ge 7$ constraints as $(14, 8)$. The ratio of red to blue marbles from this minimal pair is $14:8$, simplifying to $7:4$.

**Finalizing Marble Ratio**

My thoughts are now solidified. I've confirmed the core equation $3R - 5B = 2$ and rigorously verified that the smallest integer solution pair satisfying the constraints $R \ge 7$ and $B \ge 7$ is $(14, 8)$. This pair directly yields the ratio of red to blue marbles as $14:8$, which simplifies to $7:4$.
\end{lstlisting}

\captionof{reasoningprocess}{Correct Reasoning Process of AIME26 Problem 20}
\label{reasonaime26p20correct}
\begin{lstlisting}[style=promptbox]
The problem asks us to find the sum of the five least values of $n$ for which the following conditions hold:
1.  An urn contains $n$ marbles, each either red or blue.
2.  There are at least 7 marbles of each color. Let $R$ be the number of red marbles and $B$ be the number of blue marbles. Thus, $R \ge 7$ and $B \ge 7$. This implies $n = R+B \ge 7+7=14$.
3.  When 7 marbles are drawn randomly without replacement, the probability of drawing exactly 4 red marbles equals the probability of drawing exactly 5 red marbles.

The probability of drawing exactly $k$ red marbles in a sample of size $m$ from an urn with $R$ red and $B$ blue marbles (total $n=R+B$) is given by the hypergeometric distribution formula:
$P(X=k) = \frac{\binom{R}{k} \binom{B}{m-k}}{\binom{n}{m}}$

In this problem, $m=7$. The condition is $P(X=4) = P(X=5)$.
$\frac{\binom{R}{4} \binom{B}{7-4}}{\binom{n}{7}} = \frac{\binom{R}{5} \binom{B}{7-5}}{\binom{n}{7}}$

Since $\binom{n}{7}$ is the same on both sides (and non-zero as $n \ge 14$), we can simplify this to:
$\binom{R}{4} \binom{B}{3} = \binom{R}{5} \binom{B}{2}$

We can rewrite the binomial coefficients using the identity $\binom{n}{k} = \frac{n}{k} \binom{n-1}{k-1}$ or by expanding them.
Using the ratio form: $\frac{\binom{n}{k}}{\binom{n}{j}} = \frac{n!/(k!(n-k)!)}{n!/(j!(n-j)!)} = \frac{j!(n-j)!}{k!(n-k)!}$.
Let's rearrange the equation:
$\frac{\binom{R}{4}}{\binom{R}{5}} = \frac{\binom{B}{2}}{\binom{B}{3}}$

Now, let's evaluate the ratios:
$\frac{\binom{R}{4}}{\binom{R}{5}} = \frac{R!/(4!(R-4)!)}{R!/(5!(R-5)!)} = \frac{5!(R-5)!}{4!(R-4)!} = \frac{5 \cdot 4! \cdot (R-5)!}{4! \cdot (R-4) \cdot (R-5)!} = \frac{5}{R-4}$ (Requires $R \ge 5$)

$\frac{\binom{B}{2}}{\binom{B}{3}} = \frac{B!/(2!(B-2)!)}{B!/(3!(B-3)!)} = \frac{3!(B-3)!}{2!(B-2)!} = \frac{3 \cdot 2! \cdot (B-3)!}{2! \cdot (B-2) \cdot (B-3)!} = \frac{3}{B-2}$ (Requires $B \ge 3$)

The equality becomes:
$\frac{5}{R-4} = \frac{3}{B-2}$

Cross-multiplying gives:
$5(B-2) = 3(R-4)$
$5B - 10 = 3R - 12$
$5B + 2 = 3R$

Now, we use the fact that $n = R+B$, so $B = n-R$. Substitute this into the equation:
$5(n-R) + 2 = 3R$
$5n - 5R + 2 = 3R$
$5n + 2 = 8R$
$R = \frac{5n+2}{8}$

Now we find $B$ in terms of $n$:
$B = n - R = n - \frac{5n+2}{8} = \frac{8n - (5n+2)}{8} = \frac{3n-2}{8}$

For $R$ and $B$ to be integers, $5n+2$ and $3n-2$ must be divisible by 8.
If $5n+2$ is divisible by 8, then $5n+2 \equiv 0 \pmod{8}$.
$5n \equiv -2 \equiv 6 \pmod{8}$.
Multiplying by 5 (which is the multiplicative inverse of 5 modulo 8, since $5 \times 5 = 25 \equiv 1 \pmod{8}$):
$5 \times 5n \equiv 5 \times 6 \pmod{8}$
$25n \equiv 30 \pmod{8}$
$n \equiv 6 \pmod{8}$

Let's check if $B = \frac{3n-2}{8}$ is also an integer when $n \equiv 6 \pmod{8}$.
If $n = 8k+6$ for some integer $k$:
$3n-2 = 3(8k+6) - 2 = 24k + 18 - 2 = 24k + 16 = 8(3k+2)$. This is divisible by 8.
So, the condition $n \equiv 6 \pmod{8}$ ensures that both $R$ and $B$ are integers.

Now we apply the constraints $R \ge 7$ and $B \ge 7$:
1.  $R \ge 7$:
    $\frac{5n+2}{8} \ge 7$
    $5n+2 \ge 56$
    $5n \ge 54$
    $n \ge \frac{54}{5} = 10.8$

2.  $B \ge 7$:
    $\frac{3n-2}{8} \ge 7$
    $3n-2 \ge 56$
    $3n \ge 58$
    $n \ge \frac{58}{3} = 19.333\dots$

We need $n$ to satisfy $n \equiv 6 \pmod{8}$ and $n \ge 19.333\dots$.
Let's list values of $n$ that satisfy $n \equiv 6 \pmod{8}$:
$6, 14, 22, 30, 38, 46, 54, 62, \dots$

The smallest value of $n$ that is greater than or equal to $19.333\dots$ and satisfies $n \equiv 6 \pmod{8}$ is $n=22$.
The five least values of $n$ are obtained by starting from $n=22$ and adding the common difference of 8:
-   $n_1 = 22$
-   $n_2 = 22 + 8 = 30$
-   $n_3 = 30 + 8 = 38$
-   $n_4 = 38 + 8 = 46$
-   $n_5 = 46 + 8 = 54$

We are asked to find the sum of these five least values of $n$.
Sum $= 22 + 30 + 38 + 46 + 54$
This is an arithmetic series with $a_1 = 22$, $d=8$, and $k=5$ terms.
Sum $= \frac{k}{2}(2a_1 + (k-1)d) = \frac{5}{2}(2 \times 22 + (5-1) \times 8) = \frac{5}{2}(44 + 4 \times 8) = \frac{5}{2}(44 + 32) = \frac{5}{2}(76) = 5 \times 38 = 190$.

The final answer is $\boxed{190}$.
\end{lstlisting}

\captionof{reasoningprocess}{Incorrect Reasoning Process of GPQA Diamond Problem 197}
\label{reasongpqadiamondp197incorrect}
\begin{lstlisting}[style=promptbox]
Compound X is identified from the IR and 1H NMR data.
The IR spectrum shows characteristic absorptions for a carboxylic acid (broad O-H stretch at 3400-2500 cm-1 and C=O stretch at 1720 cm-1) and an aromatic ring (C=C stretches at 1610 cm-1 and 1450 cm-1).
The 1H NMR spectrum shows:
- A broad singlet at 10.5 ppm (1H) for the carboxylic acid proton (-COOH).
- Two doublets in the aromatic region at 8.0 ppm (2H) and 7.2 ppm (2H), indicating a para-substituted benzene ring. The downfield shift of the 8.0 ppm protons suggests they are ortho to an electron-withdrawing group.
- Aliphatic signals: 2.9 ppm (m, 1H), 1.7 ppm (m, 2H), 1.4 ppm (d, 3H), 0.9 ppm (t, 3H).
The combination of a triplet at 0.9 ppm (3H) and a multiplet at 1.7 ppm (2H) is characteristic of an ethyl group (-CH2CH3).
The combination of a doublet at 1.4 ppm (3H) and a multiplet at 2.9 ppm (1H) is characteristic of a methine group with an adjacent methyl group (-CH(CH3)-).
Putting these together with the para-substituted aromatic ring and the carboxylic acid group, the structure of compound X is identified as 4-(sec-butyl)benzoic acid. The sec-butyl group is -CH(CH3)CH2CH3, where the benzylic proton is at 2.9 ppm, the methyl attached to it is at 1.4 ppm, the methylene is at 1.7 ppm, and the terminal methyl is at 0.9 ppm.

The reaction of carboxylic acids with red phosphorus and hydroiodic acid (HI) is a strong deoxygenation process that typically converts the carboxylic acid group (-COOH) into a methyl group (-CH3), effectively reducing it to an alkane. For an aromatic carboxylic acid like benzoic acid, this reaction converts it to toluene (methylbenzene).
Therefore, 4-(sec-butyl)benzoic acid would be expected to react with red phosphorus and HI to form 4-(sec-butyl)benzene.

Let's examine the given options:
A. 1-(sec-butyl)-4-methylbenzene
B. 1-isobutyl-4-methylbenzene
C. 4-(sec-butyl)benzoic acid (This is compound X itself, not the product of the reaction).
D. 2-(4-ethylphenyl)propanoic acid

Our predicted product, 4-(sec-butyl)benzene, is not among the options. This suggests that either there is a mistake in the question or the options provided, or the reaction proceeds via a more complex pathway, possibly involving rearrangement or fragmentation of the sec-butyl group and transformation of the carboxylic acid.

Given the standard reactivity of carboxylic acids with HI/P, none of the options directly corresponds to the expected product. However, if we are forced to choose among the options and assume a complex rearrangement occurs, option D represents a significant structural change that involves both rearrangement of the alkyl group and transformation of the functional group. Without a known reaction pathway for this transformation under these conditions, a rigorous step-by-step derivation to option D is not possible.

Assuming there is an intended correct answer among the options, and given the complexity of the provided options, it is possible that the reaction involves an intricate rearrangement and functionalization of the sec-butyl group and the carboxylic acid moiety. However, based on established organic chemistry principles, the reaction of 4-(sec-butyl)benzoic acid with red phosphorus and HI should yield 4-(sec-butyl)benzene. Since this is not an option, and we cannot justify any of the given options through standard reaction mechanisms, there might be an error in the problem statement or the provided options.

However, if this is a graduate-level problem, it might involve a non-trivial reaction. Lacking specific knowledge of such a reaction, I cannot provide a definitive step-by-step derivation to option D. Therefore, I must conclude that the problem as stated does not allow for a clear solution among the given choices based on standard organic chemistry reactions.

The final answer is $\boxed{D}$.
\end{lstlisting}

\captionof{reasoningprocess}{Correct Reasoning Process of GPQA Diamond Problem 197}
\label{reasongpqadiamondp197correct}
\begin{lstlisting}[style=promptbox]
**1. Analyze the Question and Data**

The problem asks us to identify the final product of a reaction between an unknown Compound X and red phosphorus (P) and hydroiodic acid (HI), given the IR and ^1H NMR spectra of Compound X.

**2. Deduce the Structure of Compound X from Spectroscopic Data**

*   **IR Spectroscopy:**
    * The broad absorption in the range of 3400-2500 cm^{-1} is characteristic of the O-H stretching vibration of a carboxylic acid, broadened by hydrogen bonding.
    *   The strong absorption at 1720 cm^{-1} is indicative of a C=O stretching vibration, consistent with a carboxylic acid carbonyl group.
    *   The bands at 1610 cm^{-1} and 1450 cm^{-1} suggest the presence of an aromatic ring (C=C stretching and C-H bending vibrations).
    *   **Conclusion from IR:** Compound X is an aromatic carboxylic acid.

*   **^1H NMR Spectroscopy:**
    *   **10.5 ppm (bs, 1H):** This signal is a broad singlet integrating for one proton, highly deshielded. This is characteristic of the acidic proton of a carboxylic acid (-COOH).
    *   **8.0 ppm (d, 2H) and 7.2 ppm (d, 2H):** These signals are in the aromatic region. The presence of two doublets, each integrating for two protons, with characteristic chemical shifts, strongly suggests a *para*-disubstituted benzene ring. The protons at 8.0 ppm are ortho to a more electron-withdrawing group (likely the carboxylic acid), and the protons at 7.2 ppm are ortho to a less electron-withdrawing or electron-donating group (an alkyl substituent).
    *   **Aliphatic Region (2.9 ppm (m, 1H), 1.7 ppm (m, 2H), 1.4 ppm (d, 3H), 0.9 ppm (t, 3H)):** These signals represent the protons of an alkyl substituent attached to the aromatic ring. Let's analyze them:
        *   The triplet at 0.9 ppm (3H) is characteristic of a terminal methyl group (CH_3) adjacent to a methylene group (CH_2).
        *   The doublet at 1.4 ppm (3H) is characteristic of a methyl group (CH_3) adjacent to a methine (CH) group.
        *   The multiplet at 1.7 ppm (2H) is likely a methylene group (CH_2) adjacent to both the terminal CH_3 and the CH.
        *   The multiplet at 2.9 ppm (1H) is likely a methine (CH) group attached to the benzene ring, adjacent to the CH_2 and the methyl group.
        *   This pattern of CH_3-CH_2-CH(CH_3)- is characteristic of a **sec-butyl group**.
    *   **Conclusion from ^1H NMR:** Compound X is a *para*-disubstituted benzene ring, with one substituent being a carboxylic acid and the other being a sec-butyl group.

*   **Putting it together:** The structure of Compound X is **4-(sec-butyl)benzoic acid**.

**3. Predict the Product of the Reaction**

Compound X (4-(sec-butyl)benzoic acid) reacts with red phosphorus (P) and hydroiodic acid (HI). This combination of reagents is a powerful reducing system used to convert carboxylic acids into primary alkanes. The reaction mechanism involves the reduction of the carboxylic acid group (-COOH) to a methyl group (-CH_3).

The reaction is:
R-COOH + P/HI rightarrow R-CH_3

Applying this to our Compound X:
4-(sec-butyl)phenyl-COOH + P/HI rightarow 4-(sec-butyl)phenyl-CH_3

The sec-butyl group remains intact. The carboxylic acid functional group (-COOH) is replaced by a methyl group (-CH_3).
The resulting structure is a benzene ring substituted at the *para* positions with a methyl group and a sec-butyl group.

This compound can be named as:
*   **4-(sec-butyl)toluene** (using toluene as the parent name)
*   **1-methyl-4-(sec-butyl)benzene** (using benzene as the parent name, numbering methyl as 1 and sec-butyl as 4)
*   **1-(sec-butyl)-4-methylbenzene** (using benzene as the parent name, numbering sec-butyl as 1 and methyl as 4, which is also a valid IUPAC naming convention where substituents are ordered alphabetically, sec-butyl before methyl, and assigned lowest locants).

**4. Evaluate the Options**

Let's compare our predicted product with the given options:

*   **A. 2-(4-ethylphenyl)propanoic acid:** This is a carboxylic acid, which is the starting material (Compound X), not the product of the reduction. Furthermore, the substituents (ethylphenyl and propanoic acid) do not match the spectroscopic data or the reaction outcome.
*   **B. 1-(sec-butyl)-4-methylbenzene:** This structure consists of a benzene ring with a methyl group and a sec-butyl group in *para* positions. This exactly matches our predicted product.
*   **C. 4-(sec-butyl)benzoic acid:** This is Compound X itself, the starting material, not the final product of the reaction with P/HI.
*   **D. 1-isobutyl-4-methylbenzene:** This compound has an isobutyl group ((CH_3)_2CH-CH_2-) instead of a sec-butyl group. The ^1H NMR data for Compound X clearly indicated a sec-butyl group (triplet at 0.9 ppm and doublet at 1.4 ppm), ruling out an isobutyl group (which would have a doublet at 0.9 ppm for two methyls and a doublet at 2.5 ppm for the CH_2).

**5. Conclusion**

Based on the spectroscopic data, Compound X is 4-(sec-butyl)benzoic acid. The reaction with red phosphorus and hydroiodic acid reduces the carboxylic acid group to a methyl group. The resulting product is 1-methyl-4-(sec-butyl)benzene (or 4-(sec-butyl)toluene), which is listed as option B.

**Relevant Insights Used:**
*   **insight_problem_framing_and_goal_definition_1**: Clearly defining the problem as identifying the product from spectroscopic data and reaction conditions.
*   **insight_mechanism_principle_congruence_analysis_1**: Used to interpret IR and NMR data by mapping spectral features to structural principles (functional groups, substitution patterns, alkyl chain structures) and to understand the reaction mechanism of R-COOH reduction to R-CH_3.
*   **insight_goal_oriented_answer_selection_prioritization_1**: Used to select the most accurate answer by prioritizing the structure that perfectly matches the deduced starting material and the predicted reaction product, and by eliminating options that do not fit the data or the reaction outcome.

The final answer is $\boxed{B}$.
\end{lstlisting}
}

\subsection{Example Reasoning Traces in \Cref{sec:insight}}
\label[appendix]{appx:completereasoningreasearch}
\captionof{reasoningtrace}{Inverse Frequency Balancing}
\begin{lstlisting}[style=promptbox]
    This method flattens long-tailed (Zipfian) data distributions to prevent models from overfitting to common 'head' concepts while underfitting rare 'tail' concepts. It uses a probabilistic sampling approach based on a maximum frequency threshold to ensure a more uniform distribution of concepts. When to use: When training foundation models or classifiers on natural data distributions where a small number of classes or concepts dominate the dataset (e.g., millions of 'cat' images vs. few 'specific species'), and the goal is to maximize zero-shot generalization or task-agnostic performance. Step-by-step: 1) Scan the dataset to calculate the total count ($N$) for every specific metadata concept or class 2) Define a saturation threshold ($t$) that represents the desired maximum number of samples per concept (e.g., $t=20,000$ was used for 400M total items) 3) Compute a sampling probability for each concept using the formula $P = \\min(1, t/N)$, ensuring tail concepts are kept with probability 1 and head concepts are downsampled 4) Iterate through the data points again; for items associated with multiple concepts, generate a random number against the probability of each associated concept 5) Select the data point for the final training set if *at least one* of its associated concepts triggers a selection, thereby preserving diversity even within common items.
\end{lstlisting}

\captionof{reasoningtrace}{Directional Skill Evolution}
\begin{lstlisting}[style=promptbox]
This technique involves systematically transforming specific, verified solutions into more general and reusable tools (skills) to enrich a knowledge base. Instead of just storing a specific answer, the system actively 'evolves' it to cover broader cases. When to use: When building a knowledge base or skill library from specific examples, particularly in domains like mathematics or coding where a specific solution (e.g., for 2D vectors) often hints at a general principle (e.g., for N-dimensions). It is used to proactively expand the capabilities of an AI agent without requiring new external problem sets. Step-by-step: 1) Select a specific, verified skill or lemma from the existing library (the seed) 2) Choose one of four specific evolution heuristics: 'Parameterize' (replace specific constants like numbers with variables), 'Extend Dimensions' (generalize from specific dimensions to n-dimensions), 'Scale Complexity' (create a more complex version of the problem), or 'Identify Key Concepts' (extract and formalize the core underlying principle) 3) Retrieve relevant unsolved problems or requests to guide the evolution context, ensuring the new skill aims at actual gaps 4) Prompt the model to rewrite the seed skill according to the chosen heuristic 5) Formally verify the generated skill; if valid and distinct from existing skills, add it to the library.
\end{lstlisting}

\end{document}